\documentclass{ceurart}

\usepackage{amsmath}
\usepackage{graphicx}
\usepackage{microtype}

\usepackage{amsthm}
\newtheorem{theorem}{Theorem}

\newtheorem{definition}{Definition}
\newtheorem{example}{Example}
\newtheorem{proposition}{Proposition}
\newtheorem{corollary}{Corollary}

\begin{document}


\newcommand{\lang}{\ensuremath{\mathcal{L}}}
\newcommand{\KB}{\ensuremath{\mathcal{K}}} 
\newcommand{\Mod}{\ensuremath{\textbf{Mod}}}
\newcommand{\ModK}{\ensuremath{\Mod(\KB)}}
\newcommand{\sem}{\ensuremath{\mathcal{S}}}

\newcommand{\CN}{\ensuremath{{\sf N}_{\mathscr{C}}}} 
\newcommand{\RN}{\ensuremath{{\sf N}_{\mathscr{R}}}} 
\newcommand{\IN}{\ensuremath{{\sf N}_{\mathscr{I}}}} 
\newcommand{\Names}{\ensuremath{{\sf N}}}  
\newcommand{\dlAnd}{\sqcap}
\newcommand{\dlOr}{\sqcup}
\newcommand{\subs}{\sqsubseteq}
\newcommand{\nsubs}{\not\subs}
\newcommand{\I}{\ensuremath{\mathcal{I}}}
\newcommand{\J}{\ensuremath{\mathcal{J}}}
\newcommand{\Iof}[1]{{#1}^{\I}}
\newcommand{\Db}{\mathcal{D}}
\newcommand{\Dom}{\Delta}
\newcommand{\Ifunc}{\Iof{\cdot}}
\newcommand{\ALC}{\ensuremath{\mathcal{ALC}}}
\newcommand{\ALCO}{\ensuremath{\mathcal{ALCO}}}
\newcommand{\SHIF}{\ensuremath{\mathcal{SHIF}}}
\newcommand{\SHOIN}{\ensuremath{\mathcal{SHOIN}}}
\newcommand{\SHOIQ}{\ensuremath{\mathcal{SHOIQ}}}
\newcommand{\SROIQ}{\ensuremath{\mathcal{SROIQ}}}
\newcommand{\SHOQ}{\ensuremath{\mathcal{SHOQ}}}
\newcommand{\EL}{\ensuremath{\mathcal{EL}}}
\newcommand{\inv}{^{-}} 
\newcommand{\ABox}{\ensuremath{\mathcal{A}}}
\newcommand{\TBox}{\ensuremath{\mathcal{T}}}
\newcommand{\Ont}{\ensuremath{\mathcal{O}}}

\newcommand{\C}{\ensuremath{\mathcal{C}}}
\newcommand{\T}{\ensuremath{\mathcal{T}}}
\newcommand{\D}{\ensuremath{\mathcal{D}}}
\newcommand{\A}{\ensuremath{\mathcal{A}}}

\newcommand{\NC}{\ensuremath{\mathcal{N}}_{C}}
\newcommand{\NI}{\ensuremath{\mathcal{N}}_{I}}
\newcommand{\NR}{\ensuremath{\mathcal{N}}_{R}}
\newcommand{\N}{\mathcal{N}}


\newcommand{\dentails}{\mid\hskip-0.40ex\approx}   
\newcommand{\ndentails}{\not\mid\hskip-0.40ex\approx}

\newcommand{\twiddle}{\mathrel|\joinrel\sim}        
\newcommand{\ntwiddle}{\mathrel|\joinrel\not\sim}   

\newcommand{\dsubsumes}{\:\raisebox{0.45ex}{\ensuremath{\sqsubset}}\hskip-1.7ex\raisebox{-0.6ex}{\scalebox{0.9}{\ensuremath{\sim}}}\:}
\newcommand{\ndsumsumes}{\not\hspace{-1.1mm}\dsubsumes}

\newcommand{\nicholas}[1]{{\leavevmode \color{red} NL: }{\leavevmode \color{blue}#1}}
\newcommand{\giovanni}[1]{{\leavevmode \color{red} GC: }{\leavevmode \color{blue}#1}}
\newcommand{\tommie}[1]{{\leavevmode \color{red} TM: }{\leavevmode \color{blue}#1}}

\newcommand{\tuple}[1]{\langle #1 \rangle}
\newcommand{\sat}{\Vdash}

\newcommand{\cass}[2]{\mbox{$#1$:\hspace{0.1cm}$#2$}}
\newcommand{\rass}[3]{\mbox{$(#1,#2)$:\hspace{0.1cm}$#3$}}
\newcommand{\dsubs}{\dsubsumes}

\newcommand{\U}{\mathcal{U}}

\newcommand{\Nick}[1]{\textcolor{blue}{#1}}

\title{Semantic Bridges Between First Order $c$-Representations and Cost-Based Semantics: An Initial Perspective}

\author[1]{Nicholas Leisegang}[%
orcid=0000-0002-8436-552X,
email=lsgnic001@myuct.ac.za,
]
\author[2,1]{Giovanni Casini}[orcid=0000-0002-4267-4447, email=giovanni.casini@isti.cnr.it]
\author[1]{Thomas Meyer}[orcid=0000-0003-2204-6969, email=tmeyer@airu.co.za]

\address[1]{University of Cape Town and CAIR,
  Rondebosch, Cape Town, 7700, South Africa}
\address[2]{CNR - ISTI, Pisa, Italy}

\begin{abstract}
Weighted-knowledge bases and cost-based semantics represent a recent formalism introduced  by Bienvenu et al. for Ontology Mediated Data Querying in the case where a given knowledge base is inconsistent. 
This is done by adding a weight to each statement in the knowledge base (KB), and then giving each DL interpretation a cost based on how often it breaks rules in the KB. In this paper we compare this approach with \emph{$c$-representations}, a form of non-monotonic reasoning originally introduced by Kern-Isberner. 
$c$-Representations describe a means to interpret defeasible concept inclusions of the form $C\dsubs D$ (read ``instances of $C$ are usually instances of $D$'') in the first-order case. This is done by assigning a numerical ranking to each interpretations via penalties for each violated conditional. We compare these two approaches on a semantic level. In particular, we show that under certain conditions a weighted knowledge base and a set of defeasible conditionals can generate the same ordering on interpretations, and therefore an equivalence of semantic structures up to relative cost. Moreover, we compare entailment described in both cases, where certain notions are equivalently expressible in both formalisms. Our results have the potential to benefit further work on both cost-based semantics and c-representations. 
\end{abstract}

\conference{
}
\maketitle

\section{Introduction}

Description logics~(DLs)~\cite{BaaderEtAl2007} provide the logical foundation for formal ontologies of the OWL family~\cite{OWL2}.
Many  extensions of classical DLs have been proposed to enrich the representational capabilities of DLs, especially to support reasoning under forms of uncertainty. Two aspects of uncertain reasoning that have attracted the attention of the community are defeasible reasoning \cite{BaaderHollunder1995,BonattiEtAl2015,Bonatti09a,CasiniStraccia13,DoniniEtAl2002,Giordano13,GiordanoEtAl2015} and inconsistency handling \cite{LemboEtAl2010,BienvenuEtAl2014,BienvenuEtAl2016,Pena-SSWS20}.

In this paper we consider two specific semantic constructions, each connected to one of these two areas. On the side of inconsistency handling, we take under consideration a recent proposal by Bienvenu et al. \cite{BienvenuJeanCostBased}, formulated for the DL \ALCO. In the area of defeasible reasoning, we will consider $c$-representations \cite{KernIsberner2001}, a semantic framework based on ranking functions, and we refer to its formulation for First Order Logic (FOL) \cite{DBLP:conf/nmr/HahnKM24}, that we constrain to the expressivity of \ALCO. Both these semantic constructions are based on a similar idea: ranking the interpretations according to a numeric value that is determined by the amount of information that each interpretation violates.

We present a formal comparison between these two frameworks. The main contribution of the paper is to show that, under certain conditions, both semantic structures are equivalent up to relative cost. That is, on one hand we prove that for any ranking function defining a $c$-representation of a defeasible $\ALCO$ knowledge base $\KB$, we can construct a weighted knowledge base $\KB_\omega$ that induces the same ordering of interpretations, according to the methods in~\cite{BienvenuJeanCostBased}. On the other hand, we provide necessary and sufficient conditions that a weighted knowledge base must satisfy for the converse to hold. We also show how certain entailment relations defined in each semantic framework can be equivalently expressed in terms of the other.

The paper is organised as follows: Sections \ref{sect_costsemantics} and \ref{sect_c_repres} present  the two formal frameworks we refer to: \emph{cost-based semantics} and \emph{c-representations}, respectively; in Section \ref{sect_comparison_main} we present the focus of the paper, making a first comparison between the semantic structures and the forms of reasoning that we can model in these two structures; finally, in Sections \ref{sect_related_work} and \ref{sect_conclusion} we mention related and future work. 

\section{Background}\label{sect_back}

\subsection{Description Logic \ALCO}\label{sect_DL}

In this section we provide a brief introduction to the description logic \ALCO \cite{BaaderEtAl2007}.
Let \(\NC, \NR, \NI\) 
be finite, disjoint sets of symbols. These symbols will be our \emph{concept names}, \emph{role names},  and \emph{individual names},
respectively, that together define a \emph{vocabulary}.
Based on this we define the concept language $\lang_{C}$ of \ALCO, where a concept $C\in\lang_{C}$ iff
\[
C ::= A \mid \{o\} \mid \neg C \mid C \sqcap D \mid C \sqcup D \mid \exists r.C \mid \forall r.C \mid \bot \mid \top
\]
 for \(A \in \NC, r \in \NR, o\in\NI\), and \(C, D \in \lang_{C}\).
An ABox \(\ABox\) is a finite set of \emph{assertions} of the form $\cass{a}{A}$ and $\cass{(a,b)}{r}$ where $a,b\in\NI$, $A\in\NC$ and $r\in\NR$. These are called concept and role assertions respectively.
%
A TBox \(\TBox\) is a finite set of \emph{concept inclusions} (GCIs) of the form $C \subs D$ 
for \(C, D \in \lang_{C}\). An $\ALCO$ \emph{knowledge base} is a pair $\KB=(\T,\A)$ where $\T$ is a TBox and $\A$ is an ABox.

Interpretations give a semantics to assertions and 
 concept inclusions.
An interpretation is a tuple \(\I = \tuple{\Delta^{\I}, \cdot^\I}\) where \(\Delta^{\I}\) is a non-empty set called the \emph{domain} and \(\cdot^\I\) is a function that maps every \(a \in \NI\) to an individual \(a^\I \in \Delta^{\I}\), 
every 
\(r \in \NR\) to a set of ordered pairs \(r^\I \subseteq \Delta^{\I}\times \Delta^{\I}\), and every \(A \in \NC\) to a set \(A^\I \subseteq \Delta^{\I}\).
Interpretations are lifted to complex concepts by:

\begin{gather*}
\bot^\I = \emptyset, \qquad\quad \thickspace \top^\I = \Delta^{\I}, \qquad\quad  \{o\}^\I=o^\I, \qquad\quad  (\neg C)^\I = \Delta^{\I} \setminus C^\I, \\
(C \sqcap D)^\I = C^\I \cap D^\I, \qquad\quad  (C \sqcup D)^\I = C^\I \cup D^\I,\\
(\exists r.C)^\I = \{x \in \Delta^{\I} \mid \text{there is an \((x, y) \in r^\I\) \text{ s.t. }\(y \in C^\I\)}\},\\
(\forall r.C)^\I = \{x \in \Delta^{\I} \mid y \in C^\I \text{ for all }
(x, y) \in r^\I\}.
\end{gather*}

An interpretation \(\I\) is a \emph{model} of a concept inclusion \(C \subs D\) if $C^\I\subseteq D^\I$.
\(\I\) is a model of the assertion \(\cass{a}{A}\) if \(a^\I \in A^\I\), and \(\I\) is a model of \(\cass{(a,b)}{r}\) if \((a^\I, b^\I) \in r^\I\). This is denoted $\I\sat \tau$ for any assertion or GCI $\tau$. $\I$ is a model of the knowledge base $\KB=(\T,\A)$ (denoted $\I\Vdash \KB$) if $\I\sat \tau$ for all $\tau\in\T\cup\A$.
A knowledge base $\KB$ \emph{entails} a concept inclusion or assertion $\tau$ if $\I\Vdash \KB$ implies $\I\Vdash \tau$.

 Concept inclusions \(C \subs D\), read as ``Every $C$ is a $D$'',  represent the main element of a DL knowledge base, but they do not allow for exceptions. To do so, \emph{defeasible concept inclusions} (DCIs) $C \dsubs D$ have been introduced in various forms \cite{BritzEtAl21,GiordanoEtAl2015,BonattiEtAl2015} and associated to different semantics, but always with the same intended reading: $C\dsubs D$ stands for ``Typically, an element of $C$ is also an element of $D$''. In Section \ref{sect_c_repres} we provide an interpretation of DCIs by adapting to $\ALCO$ a semantic approach defined for defeasible conditionals in FOL.

\subsection{Cost-Based Semantics}\label{sect_costsemantics}

Cost-based semantics in DLs were introduced by Bienvenu et al. \cite{BienvenuJeanCostBased} in order to facilitate query answering over inconsistent DL knowledge bases. Intuitively, they assign a numerical ``penalty'' to each element of an ABox and TBox in the knowledge base. Then an interpretation $\mathcal{I}$ is assigned a weight based on how many elements of $\Delta^\mathcal{I}$ ``break'' rules of the knowledge base. This is defined as follows.

\begin{definition} [\cite{BienvenuJeanCostBased}]
    A \emph{weighted knowledge base (WKB)} is a pair $\KB_\omega=((\TBox,\ABox),\omega)$ where $(\TBox,\ABox)$ is a classical DL KB and $\omega:\TBox\cup\ABox\rightarrow \mathbb{N}\cup\{\infty\}$ is a cost function.
\end{definition}


This cost function is then migrated from the knowledge base to an interpretation by tracking the \textit{violations} of each rule in the interpretation. 

\begin{definition}[\cite{BienvenuJeanCostBased}]
    The set of \emph{violations of a GCI}, $C\subs D$, is given by,
    \[\textit{vio}_{C\subs D}(\I)=(C\dlAnd\neg D)^{\I}\]

    while the \emph{violations of an ABox} $\ABox$ in $\I$ are given by 
      \[\textit{vio}_{\ABox}(\I)=\{\alpha\in\ABox\mid \I\nvDash \alpha\}\]
\end{definition}

\begin{definition}[\cite{BienvenuJeanCostBased}]\label{def_cost_world}
    Given a WKB $\KB_\omega=((\TBox,\ABox),\omega)$, the \emph{cost} of an interpretation $\I$ with regards to $\KB_\omega $ is given by,

\begin{displaymath}
    \textit{cost}_{\KB_\omega}(\I)=\sum\limits_{\tau\in\TBox}\omega(\tau)|\textit{vio}_{\tau}(\I)|+\sum\limits_{\alpha\in\textit{vio}_\ABox(\I)}\omega(\alpha),
\end{displaymath}

where $|\textit{vio}_{\tau}(\I)|$ is the cardinality of the set $\textit{vio}_{\tau}(\I)$. We will, in some cases, omit the subscript from the function $\textit{cost}_{\KB_\omega}$. For each WKB $\KB_\omega=((\TBox,\ABox),\omega)$ we can define the \emph{optimal cost} of $\KB_\omega$ as 
    $\textit{optc}(\KB_\omega)=\min_\I(\textit{cost}_{\KB_\omega}(\I))$.
\end{definition}


The intuition is that an interpretation with a higher cost is ``worse'' in the sense that it contradicts more axioms with a greater cost. In particular, if some TBox or ABox axiom has a cost of $\infty$, for any interpretation $\I$ breaking it we have $\textit{cost}( \I)=\infty$. From this semantics, there are several different notions of entailment which are defined \cite{BienvenuJeanCostBased}. These are originally defined for Boolean Conjunctive Queries, but it is straightforward to extend them to general $\ALCO$ statements.  

\begin{definition}[\cite{BienvenuJeanCostBased}]\label{defintion:entailments-in-cost-based-semantics}
    For any GCI or assertion $\tau$ in $\ALCO$, any weighted knowledge base $\KB_\omega$, and any $k\in\mathbb{N}$, we define the following entailment relations:
    \begin{itemize}
        \item $\KB_\omega\vDash^k_{c} \tau$ if $\I\sat \tau$ for all interpretations $\mathcal{I}$ with $\textit{cost}_{\KB_\omega}(\I)\leq k$.
        \item $\KB_\omega\vDash^k_{p} \tau$ if $\I\sat \tau$ for some interpretation $\mathcal{I}$ with $\textit{cost}_{\KB_\omega}(\I)\leq k$.
        \item $\KB_\omega\vDash^{opt}_{c} \tau$ if $\I\sat \tau$ for all interpretations $\mathcal{I}$ with $\textit{cost}_{\KB_\omega}(\I)= optc(\KB_\omega)$.
        \item $\KB_\omega\vDash^{opt}_p \tau$ if $\I\sat \tau$ for some interpretation $\mathcal{I}$ with $\textit{cost}_{\KB_\omega}(\I)= optc(\KB_\omega)$.
    \end{itemize}
\end{definition}

Note that if the knowledge base $\KB$ is classically consistent, then the optimal cost for $\KB_\omega$ is $0$. In this case, we have that $\vDash^{opt}_{c}$ coincides with classical reasoning \cite{BienvenuJeanCostBased}. Another preliminary result of particular interest in our context concerns the monotonicity of these entailment relations. For that, we first need to define monotonicity in the context of WKBs.

\begin{definition}
    Let $\vDash$ be an entailment relation defined with respect to WKBs. Then $\vDash$ is \emph{monotonic} if, for any WKB $\KB_\omega$, we have 
    
    \begin{displaymath}
        \KB_\omega\vDash \tau \implies\KB'_{\omega'}\vDash \tau
    \end{displaymath}

    where $\KB\subseteq \KB'$ and $\omega'(\tau')=\omega(\tau')$ for all $\tau'\in\KB$.
\end{definition}

This characterization of monotonicity is fairly straightforward: the knowledge base $\KB'$ extends the information contained in $\KB$. New information and its associated costs can be added to the TBox or ABox, while the existing information and its costs are preserved. Under this framework, only one of the four mentioned entailment relations is monotonic.

\begin{proposition}\label{proposition:cost-based-entailments-monotonic-or-not}
    For any $k\in\mathbb{N}$, the entailment relation $\vDash^k_{c}$ is monotonic, while $\vDash^k_{p}$, $\vDash^{opt}_{c}$ and $\vDash^{opt}_{p}$ are non-monotonic.
\end{proposition}


\subsection{Ranking Semantics and C-Representations -- From FOL to \ALCO}\label{sect_c_repres}

In this section, we present a semantic framework aimed at modeling non-monotonic reasoning: \emph{c-representations} \cite{KernIsberner2001}. They are a subclass of \emph{ranking functions}, also known as \emph{Ordinal Conditional Functions} (OCFs)\cite{Spohn2012}. Both frameworks were originally introduced for propositional logic, but extensions to a conditional language in FOL have been proposed first for ranking functions~\cite{DBLP:conf/ecai/Kern-IsbernerT12}, and subsequently tailored for c-representations~\cite{DBLP:conf/nmr/HahnKM24}. Since DLs correspond to specific fragments of FOL, these characterizations can be applied to DLs as well. Therefore, we introduce ranking functions and c-representations directly for the DL \ALCO: the following definitions are those presented in~\cite{DBLP:conf/nmr/HahnKM24} for FOL, here reformulated to match the expressivity of \emph{defeasible} \ALCO. A \emph{defeasible} \ALCO~KB is a triple $\KB=(\T,\D,\A)$, where $\T$ and $\A$ are, respectively, a TBox containing GCIs $C\subs D$, and an ABox containing assertions $\cass{a}{C}$ and $\rass{a}{b}{r}$, while $\D=\{C_1\dsubs D_1,\ldots, C_n\dsubs D_n\}$ is a finite set of DCIs \cite{BritzEtAl21}.

\vspace{-0.2cm}

\paragraph{Remark.} In order to adapt the definition for conditional FOL to defeasible \ALCO, we rely on the translation of \ALCO~statements in FOL as presented in \cite{DBLP:conf/nmr/HahnKM24}:

\begin{center}
$\begin{array}{ccc}
   \cass{a}{C}  & \rightsquigarrow & C(a) \\
    \rass{a}{b}{r} & \rightsquigarrow & r(a,b) \\
    C\subs D  & \rightsquigarrow & \forall x(C(x)\Rightarrow D(x)) \\
    C  & \rightsquigarrow & C(x) \\
    C\dsubs D  & \rightsquigarrow & (D(x)|C(x)) \\
\end{array}$
\end{center}

In particular, a defeasible inclusion is interpreted as an open conditional. We refer the reader to \cite{DBLP:conf/nmr/HahnKM24} to check the details and  notation used for conditional FOL. Here, we also define the additional symbol $\dsubs^\forall$ in order to  refer to a universally quantified defeasible conditional, as introduced in \cite{DBLP:conf/ecai/Kern-IsbernerT12}. That is:
\[C\dsubs^\forall D   \rightsquigarrow  \forall x(D(x)|C(x)). \]

Intuitively, a \emph{quantified DCI} $C \dsubs^\forall D$ expresses that, in the most typical situations, $a \in C^\I$ implies $a \in D^\I$ for \textit{every} $a \in \Delta^\I$, whereas $C \dsubs D$ expresses that, in the most typical situations, $a \in C^\I$ implies $a \in D^\I$ for \textit{certain} elements $a \in \Delta^\I$:  the most preferred (or typical) members of $C^\I$. In what follows, the technical differences between various forms of defeasible concept inclusions become apparent.

Another aspect we adapt from the FOL formulation is Herbrand semantics, a well-known form of first-order interpretations, that   in~\cite{DBLP:conf/ecai/Kern-IsbernerT12} are used to define ranking semantics for defeasible reasoning. In the present setting, we fix a finite set of individual names $\U=\NI$, referred to as the \textit{Herbrand Universe}. That is, $\U$ corresponds to the set $\NI$ in the vocabulary, hence it includes  all the individual names appearing in the KB and possibly others.  An $\ALCO$ interpretation $\I$ is a \textit{Herbrand interpretation} if $\Delta^\I = \U$ and $a^\I = a$ for all $a \in \U$. The set of all Herbrand interpretations for a given $\ALCO$ vocabulary is denoted by $\Omega_\N$. In the following, we assume that all interpretations are Herbrand interpretations defined over some vocabulary. The following definition are all reformulations of notions presented for conditional FOL in \cite{DBLP:conf/ecai/Kern-IsbernerT12,DBLP:conf/nmr/HahnKM24}.


\begin{definition}
 A \emph{ranking function} $\kappa$ is a mapping $\kappa:\Omega_\N\rightarrow \mathbb{N}\cup\{\infty\}$, such that $\kappa^{-1}(0)\neq \emptyset$.
\end{definition}

Ranking functions are generally interpreted as representing the expectations of an agent: the lower the rank associated to an interpretation, the more plausible the represented situation is considered by the agent. The interpretations with rank $0$ describe what the agent expects to hold.

Although these ranking functions are initially defined on interpretations, they can be extended to $\ALCO$ statements. 

\begin{definition}
    Let $\kappa$ be a ranking function. 

    \begin{itemize}
        \item for every assertion $\cass{a}{C}$, $\kappa(\cass{a}{C})=\min_{\I\sat a:C}\kappa(\I)$;
        \item for every assertion $\rass{a}{b}{r}$, $\kappa(\rass{a}{b}{r})=\min_{\I\sat (a,b):r}\kappa(\I)$;
        \item for every TBox statement $C\subs D$, $\kappa(C\subs D)=\min_{\I\sat C\subs D}\kappa(\I)$;
        \item for every concept $C$, $\kappa(C)=\min_{a\in\Delta^\I}\kappa(\cass{a}{C})$;

        \item for every DBox statement $C\dsubs D$, $\kappa(C\dsubs D)=\min_{a\in\Delta^\I}(\kappa(\cass{a}{C\dlAnd D})-\kappa(\cass{a}{C}))$.
    \end{itemize}

\end{definition}

Based on such ranking functions, we can define the satisfaction relation of classical \ALCO~statements and $\dsubs^\forall$-inclusions.

\begin{definition}
    Given a ranking function $\kappa$ we define satisfaction of  \ALCO~statements as follows:
    \begin{itemize}
        \item $\kappa\sat \cass{a}{C}$ iff $\I\sat \cass{a}{C}$ for all $\I\in \kappa^{-1}(0)$.
        \item $\kappa\sat \rass{a}{b}{r}$ iff $\I\sat \rass{a}{b}{r}$ for all $\I\in \kappa^{-1}(0)$.
        \item $\kappa\sat C\subs D$ iff $\I\sat C\subs D$ for all $\I\in \kappa^{-1}(0)$.
        \item $\kappa\sat C\dsubs^\forall D$ iff $\kappa(\cass{a}{C\dlAnd D})<\kappa(\cass{a}{C\dlAnd \neg D})$ for all $a\in\U$.
    \end{itemize}
\end{definition}

More elaborate is the definition of acceptance for DCIs.

\begin{definition}
    Let $C\dsubs D$ be any defeasible subsumption.  We say that $a\in\U$ is a \emph{weak representative} for $C\dsubs D$ if the following conditions hold:
    \begin{equation}
    \kappa(\cass{a}{C\dlAnd D})=\kappa(C\dlAnd D) 
    \end{equation}
    \begin{equation}
        \kappa(\cass{a}{C\dlAnd D})<\kappa(\cass{a}{C\dlAnd \neg{D}}).
    \end{equation}
and denote the set of weak representatives by $WRep(C\dsubs D)$. A \emph{strong representative } of $C\dsubs D$ is defined as a weak representative of $C\dsubs D$ such that
\begin{equation}
    \kappa(\cass{a}{C\dlAnd \neg{D}})=\min_{b\in wrep(C\scriptsize\dsubs \normalsize D)}\kappa(\cass{b}{C\dlAnd \neg{D}})
\end{equation}
    We denote the set of strong representatives by $Rep(C\dsubs D)$.
\end{definition}

The conditions above for representatives may appear somewhat technical. However, the intuition behind each condition can be described as follows. If $a$ is a (strong) representative for the rule $C \dsubs D$, then $a$ is a ``maximally typical'' instance of both $C$ and $D$ (as per condition (1)). Moreover, it is more ``typical'' for $a$ to satisfy the defeasible inclusion than to violate it (according to condition (2)), and condition (3) specifies that when $a$ does violate $C \dsubs D$, it does so for the ``most common'' reasons, compared to other groundings that satisfy (1) and (2). This intermediary definition enables us to define acceptance for open conditionals, that is, defeasible inclusions.

\begin{definition}\label{definition:satisfaction-of-open-DCIs}
    Let $\kappa$ be a ranking function and $C\dsubs D$ be a defeasible inclusion. Then we say that $\kappa$ \emph{satisfies} $C\dsubs D$ ($\kappa\sat  C\dsubs D$) iff $Rep(\tau)\neq \emptyset$ and one of the following conditions hold:
    \begin{itemize}
        \item[A.] $\kappa(C\dlAnd D)<\kappa(C\dlAnd\neg D).$
        \item[B.] $\kappa(C\dlAnd D)=\kappa(C\dlAnd\neg D)$ and either $Rep(C\dsubs\neg D)$ is empty, or for all $a\in Rep(C\dsubs D)$ and all $b\in Rep(C\dsubs\neg D)$ we have,
    $\kappa(\cass{a}{C\dlAnd\neg D})<\kappa(\cass{b}{C\dlAnd D}).$
    \end{itemize}
\end{definition}

Condition A. refers to the prototypical case where the validation of a conditional is universally more likely than a contradiction to it. Condition B. refers to the case where there are individuals on the same rank who act as a representative the rule $C\dsubs D$ and $C\dsubs \neg D$. However, in the case of B., the rule $C\dsubs D$ wins out, since its representatives violate the rule with less exceptionality. That is, the representatives of rule $C\dsubs D$ are behaving as they do not due to specific properties of their behaviour as individuals, but due to the fact that they ``fall in line'' to the general prototypical patterns of the world around them. More discussion on this can be found in \cite{DBLP:conf/ecai/Kern-IsbernerT12}. It should be noted here that the quantified DCI $C\dsubs^\forall D$ can be equivalently expressed as a set of open $\ALCO$ DCIs:

\begin{proposition}\label{proposition:express-quantified-DCI-as-normal_DCIs}
    $\kappa\Vdash C\dsubs^\forall D \text{ iff }\kappa\Vdash (\{a\}\dlAnd C)\dsubs D \text{ for all }a\in U_\Sigma$.
\end{proposition}

Hence, we are also able to include any quantified DCI $C\dsubs^\forall D$ in our knowledge base $\KB=(\T,\D,\A)$, since they can be expressed as a set of DBox axioms. 

\begin{definition}
    Let $\KB=(\T,\D,\A)$ be a knowledge base. Then a ranking function $\kappa$ is a \emph{model} of $\KB$ iff  the following conditions hold:
    \begin{enumerate}
        \item $\kappa\sat  C\dsubs D$ for all $C\dsubs D\in\D$.
        \item For all $\I\in\Omega_\N$, if\ \ $\I\not\sat \T\cup\A$ then $\kappa(\I)=\infty$.
    \end{enumerate} 
\end{definition}

Now that we have introduced ranking functions as a semantics for defeasible \ALCO~KBs, we introduce $c$-representations, a specific subclass of ranking functions. Also $c$-representations have originally been introduced for propositional logic \cite{KernIsberner2001}, and we refer to a recent reformulation for first order conditionals \cite{DBLP:conf/nmr/HahnKM24}, again, constraining it to \ALCO~expressivity. 
$c$-representations are ranking functions which assign a penalty to each defeasible conditional that occurs in the knowledge base, and then assigns a rank to a DL interpretation based on how many times the interpretation violates defeasible inclusions in the knowledge base. This is defined formally below.



\begin{definition}\label{def_c_representation}
    Let $\KB=(\T,\D,\A)$ be a defeasible $\ALCO$~KB with $\D=\{C_1\dsubs D_1,\ldots ,C_n\dsubs D_n\}$. 
A ranking function $\kappa$ is a \emph{c-representation}  of $\KB$ if it is a model 
of $\KB$ and there is some  $\{\eta_1,\ldots,\eta_n\}\subset\mathbb{N}$ and $\kappa_0\in\mathbb{Z}$ s.t., for each $\I\in\Omega_\N$ with $\I\sat  \T\cup\A$, we have  

\vspace{-0.3cm}
    \[\kappa(\I)=\kappa_0+\sum_{i=1}^{n}f_i(\I)\eta_i,\]

\vspace{-0.1cm}
   \noindent where $f_i(\I)=|\{a\in \Delta^\I\mid \I\sat  \cass{a}{C_i\dlAnd\neg D_i}\}|$. 
\end{definition}

Each value $\eta_i$ is the \textit{impact factor} for the DCI $C_i \dsubs D_i$, and its role is analogous to the weights from Section~\ref{sect_costsemantics}. Intuitively, the rank of an interpretation $\I$ under a given $c$-representation $\kappa$ is determined by counting the number of times each DCI $C_i \dsubs D_i$ is violated by $\I$, and adding the corresponding impact factor $\eta_i$ for each violation. 
The constant $\kappa_0$ serves as a \textit{normalization factor}, ensuring that $\kappa^{-1}(0) \neq \emptyset$, and thus that $\kappa$ is a valid ranking function. In particular, if there exist Herbrand interpretations that do not violate any rules, then $\kappa_0 = 0$.

It is worth noting that this is not the only semantics for DCIs in DLs. In many instances, a single DL interpretation $(\Delta^\I,\cdot^I)$ is considered, with an additional order relation $\prec\subseteq \Delta^\I\times \Delta^\I$. In this case, $\I\sat  C\dsubsumes D$ iff  $\min_\prec C^\I\subseteq D^\I$ \cite{GiordanoEtAl2015,BritzEtAl21,PenselTurhan18-defeasibleEL}. The semantic framework we use here has been shown to be sound with respect to the KLM postulates for $\dsubsumes$ developed in the $\ALC$ case\footnote{These postulates in turn are based closely on the original postulates given by Kraus et al. \cite{KrausEtAl1990}} \cite{DBLP:conf/nmr/HahnKM24}.

Looking at Definitions \ref{def_cost_world} and \ref{def_c_representation}, it is quite immediate to see that there are possible connections between the two approaches: in both cases there is a function that associates to each world a cost that is determined by the information in the KB that is falsified, and it is such a possible connection that we are going to investigate in what follows. Thus, while we acknowledge the existence of alternative semantics for defeasible DLs, in the rest of the paper we refer to the systems of defeasible reasoning based on ranking functions. When we consider logical entailment relations in the context of $c$-representations, the most basic notion is characterizing entailment using a single $c$-representation.

\begin{definition}\label{definition:kappa-entailment}
    Let $\KB=(\T,\D,\A)$ be a defeasible knowledge base,$\tau$ be an $\ALCO$ DCI, GCI, or concept or role assertion, and $\kappa$ be a $c$-representation of $\KB$. Then we say $\KB$ \emph{$\kappa$-entails} $\tau$ (written $\KB\dentails^\kappa \tau$) iff $\kappa\Vdash \tau$.
\end{definition}

This notion of entailment has been employed in other versions of ranking based semantics for defeasible  DLs \cite{DBLP:conf/nmr/HahnKM24}. One issue is to decide which $c$-representation ought to be considered for  entailment. We therefore also examine \textit{skeptical} $c$-inference \cite{BeierleEichhornKern-Isberner:skepticaljournal,BeierleEichornKern:skeptical-c-inference-article}, and \textit{credulous }$c$-inference \cite{Beierleetal:credulous-weakly-skeptical}, which are entailment relations defined considering the set of all $c$-representations. These are introduced in the propositional case with defeasible conditionals (i.e., DBox statements) specifically in mind. We extend these definitions to the DL setting, and also consider entailment for non-defeasible sentences.

\begin{definition}\label{definition:skeptical-credulous-c-inference}
     Let $\KB=(\T,\A,\D)$ be a defeasible knowledge base, and let $\tau$ be an $\ALCO$ DCI, GCI, or concept or role assertion. Then:
     \begin{enumerate}
         \item $\tau$ is a \emph{skeptical $c$-inference} of $\KB$, written $\KB\dentails^c_{sk}\tau$ iff $\KB\dentails^\kappa \tau$ for every $c$-representation $\kappa$ of $\KB$.
           \item $\tau$ is a \emph{credulous $c$-inference} of $\KB$, written $\KB\dentails^c_{cr}\tau$ iff $\KB\dentails^\kappa \tau$ for some $c$-representation $\kappa$ of $\KB$.
     \end{enumerate}
\end{definition}

Note here that while our definitions are clear extensions of the previous definitions given in \cite{BeierleEichhornKern-Isberner:skepticaljournal,BeierleEichornKern:skeptical-c-inference-article,Beierleetal:credulous-weakly-skeptical}, the syntax we give is different in order to describe the inference systems for non-defeasible statements in the language. 


\section{Comparing Ranking and Cost-Based Semantics}\label{sect_comparison_main}

In the following section we will make explicit the connection between the formalisms given in the previous two discussions. Both defeasible reasoning and cost-based semantics are methods which are introduced in DLs in order to combat contradictory information, and both take the approach that in the instance where information is contradictory, certain rules can be weakened in the knowledge base. More explicitly, both cost-based semantics and c-representations assign penalties to an interpretation based on how many elements of the interpretation break rules in the knowledge base. However, before we describe this explicitly, we must list some assumptions we make in order to compare these formalisms:

\begin{enumerate}
    \item  We only consider Herbrand interpretations, where $\U=\NI$ is a \emph{finite} Herbrand universe. This is done in order to facilitate an equivalent set of interpretations in the $c$-representation and cost-based semantics case, where $c$-representations are defined for first-order logic in terms of Herbrand semantics \cite{DBLP:conf/ecai/Kern-IsbernerT12}. It is worth noting that this does restrict the interpretations considered in the original cost-based semantics formulation given in \cite{BienvenuJeanCostBased} to a closed-world style of reasoning, where interpretations are finite. We refer to $\Delta^\I$ and $\U$ interchangeably in the rest of the section.
    \item Whenever a conditional  $C\dsubs D$ is satisfied by some $c$-representation $\kappa$, we assume that 
    $\kappa(C\dlAnd D)< \kappa(C\dlAnd \neg D)$
    and thus do not concern ourselves with  case B. in Definition \ref{definition:satisfaction-of-open-DCIs} for $\kappa$. We do this since we are interested in considering the most prototypical conditions under which a DCI is satisfied, and leave considerations of non-typical individuals for future work.  A similar assumption is used in previous literature when applying $c$-representations to DLs \cite{DBLP:conf/nmr/HahnKM24}.
\end{enumerate}

\subsection{Semantic Structures with Equivalent Relative Cost of Interpretations}\label{scet_comparison_1}

In this section, we show that for a $c$-representation $\kappa$ of some knowledge base $(\T,\D,\A)$ we can define a weighted knowledge base $\KB_{\omega}$ such that:

\[\kappa(\I)<\kappa(\I') \text{ if and only if }\textit{cost}_{\KB_\omega}(\I)<\textit{cost}_{\KB_\omega}(\I')\]

and that in certain cases the converse is possible. This shows that the way cost functions and $c$-representations structure the relative penalties attributed to interpretations may be equivalent in certain circumstances. However, the costs and the rankings of such worlds are usually not numerically equivalent, since for any $c$-representation there is always some interpretation $\I$ such that $\kappa(\I)=0$, while $\textit{cost}_{\KB_\omega}(\I)=0$ for some $\I$ only if $\KB$ is classically satisfiable. If a WKB $\KB_\omega$ and a ranking function $\kappa$ satisfy the condition given above, we say that $\KB_\omega$ and $\kappa$ have \textit{equivalent relative cost of interpretations}.

We begin by defining the derived weighted knowledge base for a given $c$-representation. 

\begin{definition}\label{definition:weight-kb-translation}
    Given a defeasible knowledge base $(\T,\D,\A)$ and a $c$-representation $\kappa$ defined by:

      \[\kappa(\I)=\kappa_0+\sum_{i=1}^{n}f_i(\I)\eta_i\]
    
    we define the \emph{weighted knowledge base translation} $\KB^\kappa_{\omega}$ as follows:
    \begin{itemize}
        \item $\KB=(\TBox^*,\ABox)$ where the ABox is the same in both cases and $\TBox^*=\T\cup \D^*$.
        \item $\D^*=\{C_i\subs D_i\mid C_i\dsubs D_i\in\D\}$.
        \item $\omega(\tau)=\infty$ for all $\tau\in \ABox\cup \TBox$.
        \item $\omega(C_i\subs D_i)=\eta_i$ for all $C_i\dsubs D_i\in\D^*$.
    \end{itemize} 
\end{definition}

It is worth immediately noting that not all WKBs can be derived from a $c$-representation using the above definition. In particular, each ABox assertion in the derived WKB is given an infinite cost and must therefore be entailed classically by every interpretation with finite cost. This results from the fact that in many defeasible reasoning formalisms in DLs there is no defeasibility included ABox statements. However, those WKBs which are derived from the above definition give us a structure with an \textit{equivalent relative cost of interpretations} to the original $c$-representation. This is formalised as follows.

\begin{proposition}\label{proposition:from-c-rep-to}
For a given $\ALC\mathcal{O}$ $c$-representation $\kappa$, and an interpretation $\I\in\Omega_\N$, we have:
    \[\textit{cost}_{\KB^\kappa_\omega}(\I)=\kappa(\I)-\kappa_0\] 
\end{proposition}

An example of such a translation can be seen below.

\begin{example}\label{example:c-representation}
    We consider an example with a structure similar to the well-known ``penguin triangle'' in non-monotonic reasoning. Let the Herbrand universe consist of a single element $\U = \{N\}$, the set of concept names be $\NC = \{\mathsf{Scientist}, \mathsf{Logician}, \mathsf{Experiments}\}$, and the set of role names be empty. Then consider the knowledge base given by:
$\A=\{\mathsf{Logician}(N)\}$, and $\D=\{\tau_1=\mathsf{Logician}\dsubs\mathsf{Scientist}, \tau_2=\mathsf{Logician}\dsubs\neg\mathsf{Experiments}, \tau_3=\mathsf{Scientist}\dsubs\mathsf{Experiments}\}$. Intuitively, this knowledge base expresses that ``logicians are usually  considered scientists,'' ``scientists usually do experiments,'' and ``logicians usually do not do experiments,'' while asserting that $N$ is a logician.

A  $c$-representation for the above knowledge base can be defined by assigning an impact factor $\eta_i$ to each $\tau_i$, where $\eta_1=1$, $\eta_2=2$ and $\eta_3=3$, and fixing $\kappa_0=-1$. Using Proposition \ref{proposition:from-c-rep-to}, we derive the WKB $\KB_\omega$ defined by:
 $\A=\{\mathsf{Logician}(N)\}$, {$\T=\{\tau'_1=\mathsf{Logician}\subs\mathsf{Scientist}, , \tau'_2=\mathsf{Logician}\subs\neg \mathsf{Experiments}, \tau'_3=\mathsf{Scientist}\subs\mathsf{Experiments}\}$}, where $\omega(\mathsf{Logician}(N))=\infty$ and $\omega(\tau'_i)=\eta_i$ for each $i$.
\end{example}

However, while we are able to construct a WKB and cost-based semantic structure from any given defeasible knowledge base and corresponding  $c$-representation, the converse is not as straightforward. This is due to the fact that, in order to translate ``weak'' TBox statements into defeasible implications we require the resulting $c$-representation to satisfy the translated knowledge base, and this is not true for all the weight-assignments. We therefore treat two different translations of weighted TBoxes into defeasible conditionals, and for now only consider those WKBs with a \textit{strict ABox}, formally defined below. We consider ``weak'' ABox rules later, using the specific expressivity in $\ALCO$ to translate weak ABox axioms into DCIs.

\begin{definition}
    A WKB $\KB_\omega$ has a \emph{strict ABox} if, for all $\tau\in \ABox$, $\omega(\tau)=\infty$.
\end{definition}

This means that any interpretation violating an ABox axiom is immediately moved to the highest infinite cost. Then, for any $\tau\in\TBox$ such that $\omega(\tau)<\infty$, we consider the following two translations:
\begin{enumerate}
    \item For any such $\tau=C\subs D$, we add $C\dsubs^\forall D$ to the defeasible knowledge base.
    \item For any such  $\tau=C\subs D$, we add $C\dsubs D$ to the defeasible knowledge base.
\end{enumerate}

 The second case, as previously mentioned, provides a translation of ``weak'' TBox statements $A\subs B$ to defeasible concept inclusions $A\dsubsumes B$, which is more faithful to the literature \cite{DBLP:conf/nmr/HahnKM24}. However, it is not always the case that such a weak concept inclusion in a weighted knowledge base is intended to represent a defeasible conditional. We therefore propose two translations, and define conditions on WKBs whose translations result in a well-defined $c$-representation.

\begin{definition}\label{definition:quantified-c-rep-translation}
    Let $\KB_\omega=((\TBox,\ABox),\omega)$ be a WKB with a  strict ABox. Then the \emph{quantified $c$-representation translation} of $\KB_\omega$ is defined as the function $\kappa_\omega$ over the knowledge base $\KB^*=(\T^\infty,\A,\D)$, where:
    \begin{itemize}
    \item The ABox is the same in both cases.
        \item $\T^\infty:=\{C\subs D\in\T\mid \omega(C\subs D)=\infty\}$ 
        \item $\D:=\{\{a\}\dlAnd C\dsubs D\mid a\in\U, C\subs D\in \T\setminus \T^\infty\}$.
       \item $\kappa_\omega(\I):=\kappa_0+\sum_{\tau\in \D} f_{\tau}(\I)\eta_\tau$ for each interpretation $\I\in\Omega_\N$ such that $\I\Vdash \T^\infty\cup \A$; $\kappa_{\omega}(\I)=\infty$ otherwise.
       \item $\eta_{\tau}:=\omega(C\subs D)$ for every $\{a\}\dlAnd C\dsubs D=\tau\in \D$.
       \item $\kappa_0:=-optc(\KB_\omega)$.
       \item $f_\tau(\I)=1$ if $\I\not\sat  \{a\}\dlAnd C\subs D$ and $f_\tau(\I)=0$ if $\I\sat  \{a\}\dlAnd C\subs D$.
    \end{itemize}
\end{definition}

\begin{definition}\label{definition:open-c-rep-translation}
    Let $\KB_\omega=((\TBox,\ABox),\omega)$ be a WKB with a  strict ABox. Then the \emph{open $c$-representation translation} of $\KB_\omega$ is defined as the function $\kappa_\omega$ over the knowledge base $\KB^*=(\T^\infty,\A,\D)$, where:
    \begin{itemize}
    \item The ABox is the same in both cases.
        \item $\T^\infty:=\{C\subs D\in\T\mid \omega(C\subs D)=\infty\}$ 
        \item $\D:=\{C\dsubs D\mid C\subs D\in \T\setminus \T^\infty\}$.
       \item $\kappa_\omega(\I):=\kappa_0+\sum_{\tau\in \D} f_{\tau}(\I)\eta_\tau$ for each interpretation $\I\in\Omega_\N$ such that $\I\Vdash \T^\infty\cup \A$; $\kappa_{\omega}(\I)=\infty$ otherwise.
       \item $\eta_{\tau}:=\omega(C\subs D)$ for every $C\dsubs D=\tau\in \D$.
       \item $\kappa_0:=-optc(\KB_\omega)$.
       \item $f_\tau(\I)=|(C\dlAnd\neg D)^\I|$ for all $\I\in \Omega_\N$ and all $\tau\in \D$.
    \end{itemize}
\end{definition}

\begin{proposition}\label{proposition:open-quantified-c-rep-translation-gives-same-ranking}
    For any WKB $\KB_\omega=((\TBox,\ABox),\omega)$, let $\kappa_1$ be the function generated by the quantified c-representation translation and let $\kappa_2$ be the function generated by the open c-representation translation. Then $\kappa_1=\kappa_2$.
\end{proposition}

\begin{proposition}\label{proposition:WKB-to-c-rep-gives-same-ranking}
    For any WKB $\KB_\omega$ with a strict ABox, we have that 
    \[\kappa_\omega(\I)=cost_{\KB_\omega}(\I)+\kappa_0\]
    for any interpretation $\I$ where $\kappa_\omega$ is the quantified or open c-representation translation of $\KB_\omega$.
\end{proposition}

The above shows that the ranking functions defined in both cases have an \textit{equivalent relative cost of interpretation} to the original WKB. Moreover, these are well-defined ranking functions, since $\kappa_0 = -optc(\KB_\omega)$ ensures that $\kappa^{-1}(0) \neq \emptyset$ in either case. However, although we refer to them as $c$-representation translations, we are not guaranteed that the resulting functions satisfy the translated knowledge base, and thus they may not be well-defined $c$-representations. We propose two conditions on WKBs that ensure their quantified and open $c$-representation translations satisfy their respective translated knowledge bases.

\begin{definition}
    A WKB with a strict ABox $\KB_\omega$ is \emph{strongly $c$-compatible} iff  for all $A\subs B\in \{\tau \in\mathcal{T}\mid \omega(\tau)< \infty\}$, $a\in\mathcal{U}$ we have 
    \begin{displaymath}
        \min_{\I\sat a: A\dlAnd B}cost_{\KB_\omega}(\I) <\min_{\I\sat a: A\dlAnd \neg B}cost_{\KB_\omega}(\I).
    \end{displaymath}
\end{definition}

This definition intuitively gives us the condition required for quantified translations to generate $c$-representations. The following gives us a definition for open translations.

\begin{definition}
    A WKB with a strict ABox $\KB_\omega$ is   \emph{$c$-compatible} iff  for all $A\subs B\in \{\tau \in\mathcal{T}\mid \omega(\tau)< \infty\}$, we have 
    \begin{displaymath}
        \min_{a\in\mathcal{U}, \I\sat a: A\dlAnd B}cost_{\KB_\omega}(\I) <\min_{a\in \mathcal{U}, \I\sat  a:A\dlAnd \neg B}cost_{\KB_\omega}(\I).
    \end{displaymath}
\end{definition}

\begin{proposition}\label{proposition:every-strong-c-compatible-is-c-compatible}
    Any strongly $c$-compatible WKB is $c$-compatible.
\end{proposition}

In order to show that these definitions are necessary, we consider the following example of a WKB which is not $c$-compatible (and therefore is also not strongly $c$-compatible):

\begin{example}
 We consider the same WKB ABox and TBox as the one derived in Example \ref{example:c-representation}. However, in this case, we change the cost function in our WKB, and consider $\KB_\omega$ defined by:
 \begin{gather*}
\omega(\mathsf{Logician}(N)) = \infty;\text{ }  \omega(\mathsf{Logician} \subs \mathsf{Scientist}) = 3;\\ \omega(\mathsf{Scientist} \subs \mathsf{Experiments}) = 2;\text{ } \omega(\mathsf{Logician} \subs \neg \mathsf{Experiments}) = 1.
\end{gather*}

We observe that for any interpretation $\I$ with cost less than 3, we must have $\I \Vdash \mathsf{Logician}(N)$ and $\I \Vdash \mathsf{Logician} \subs \mathsf{Scientist}$, which implies $\I \Vdash \mathsf{Scientist}(N)$.  
If $\I \Vdash \mathsf{Experiments}(N)$, then $(\mathsf{Logician} \dlAnd \mathsf{Experiments})^\I = \{N\}$, while $(\mathsf{Scientist} \dlAnd \neg \mathsf{Experiments})^\I = \emptyset$. Thus, $cost_{\KB_\omega}(\I) = 1$.  
On the other hand, if $\I \Vdash \neg \mathsf{Experiments}(N)$, then $(\mathsf{Logician} \dlAnd \mathsf{Experiments})^\I = \emptyset$, while $(\mathsf{Scientist} \dlAnd \neg \mathsf{Experiments})^\I = \{N\}$. In this case, $cost_{\KB_\omega}(\I) = 2$. Furthermore, due to the nature of the Herbrand interpretations considered and the fact that the given knowledge base is not classically satisfiable, these two cases represent the minimal cost for interpretations that satisfy the rules $\mathsf{Logician} \subs \neg \mathsf{Experiments}$ and $\mathsf{Scientist} \subs \mathsf{Experiments}$, respectively. Then the following holds:

\[
\min_{a \in \U, \I \Vdash a : S \dlAnd E} cost(\I) = 2 > 1 = \min_{a \in \U, \I \Vdash a : S \dlAnd \neg E} cost(\I)
\]

\noindent where $S$ and $E$ are shorthand for the predicates $\mathsf{Scientist}$ and $\mathsf{Experiments}$, respectively. Therefore, $\KB_\omega$ is not $c$-compatible when considering the rule $\mathsf{Scientist} \subs \mathsf{Experiments}$. Consequently, the quantified and open $c$-representation translations of $\KB_\omega$ do not satisfy $\mathsf{Scientist} \dsubs^\forall \mathsf{Experiments}$ and $\mathsf{Scientist} \dsubs \mathsf{Experiments}$ respectively. Moreover, Example \ref{example:c-representation} shows us that the same knowledge base $(\T,\A)$ can be $c$-compatible when considering a different cost function.
\end{example}

The following result tells us that the conditions of $c$-compatibility and strong $c$-compatibility are exactly those which allow for the open (resp. quantified) $c$-representation translations to provide us a well-defined $c$-representation.


\begin{theorem}\label{theorem:when-does-a-WKB-induce-a-c-representation}
     A WKB is strongly $c$-compatible if and only if its quantified $c$-representation translation satisfies the translated knowledge base. Similarly, a  WKB is  $c$-compatible if and only if its open $c$-representation translation satisfies the translated knowledge base.
\end{theorem}

Moreover, we are able to use open $c$-representation translations as an inverse to the weighted knowledge base translation.

\begin{theorem}\label{theorem:open-translation-inverse}
Let $\KB=(\T,\A,\D)$ be a defeasible knowledge base, and let $\KB_\omega$ be a $c$-compatible WKB. Then $\KB^{\kappa_\omega}_{\omega'}=\KB_\omega$, where $\kappa_\omega$ is the open $c$-representation translation of $\KB_\omega$, and $\kappa_{\omega''}=\kappa$, where $\kappa_{\omega''}$ is the open $c$-representation translation of $\KB^\kappa_{\omega''}$.
\end{theorem}

In our translation, we have assumed the ABox to be a strict ABox. However, in order to generalize the semantic comparison between $c$-representations and cost-based semantics, we now turn to the case of weak ABox axioms. Instead of incorporating these directly into the $c$-representation translation of a WKB, we make use of nominals in $\ALCO$ in order to translate a ``weak'' ABox axiom into a ``weak'' GCI.

\begin{definition}
    Consider a WKB $\KB_\omega=((\T,\A),\omega)$. Then the \emph{strict ABox translation} of $\KB_\omega$ is given by $\KB'_{\omega'}=((\T',\A'),\omega')$ where:
    \begin{itemize}
        \item $\A':=\{\cass{a}{A}\in\A\mid \omega(\cass{a}{A})=\infty\}$.
        \item $\T':=\T\cup\{\{a\}\subs A\mid \cass{a}{A}\in\A\setminus \A'\}$.
        \item $\omega'(\tau)=\omega(\tau)$ for all $\tau\in (\A'\cup\T)$; $\omega'(\{a\}\subs A)=\omega(\cass{a}{A})$ for all $\cass{a}{A}\in \A\setminus\A'$.     
        \end{itemize}
\end{definition}

We note below that this strict ABox translation is faithful to the original knowledge base, since it preserves costs on interpretations, and thus preserves the entailments in Definition \ref{defintion:entailments-in-cost-based-semantics}.

\begin{proposition}\label{proposition:abox-translation-preserves-costs}
    For any WKB $\KB_\omega$, we have that $cost_{\KB_\omega}(\I)=cost_{\KB'_{\omega'}}(\I)$ for all $\I\in\Omega_\N$, where $\KB'_{\omega'}$ is the strict ABox translation of $\KB_\omega$.
\end{proposition}

This in combination with our previous results shows us that any WKB whose strict ABox translation is either $c$-compatible or strongly $c$-compatible can be translated into a $c$-representation with the same relative cost of interpretations. On the other hand, every $c$-representation can be translated into a valid WKB with equivalent relative cost of interpretations.
\subsection{Bridging Entailment Relations}\label{sect_comparison_2}

In this section, we discuss the impact of the semantic comparison between cost-based semantics and $c$-representations on the entailment relations associated with each formalism. For this subsection, we assume that each WKB has a strict ABox and is strongly $c$-compatible, unless stated otherwise. We begin by comparing the entailments introduced by cost-based semantics in Definition~\ref{defintion:entailments-in-cost-based-semantics} with those defined by a single $c$-representation, as in Definition~\ref{definition:kappa-entailment}. Intuitively, these are the two most closely related entailments: in cost-based semantics, the weight of each sentence is typically treated as inherent to the knowledge base, whereas in $c$-representations, the impact factors assigned to each DCI are not intrinsic to the knowledge base but are instead part of the semantics and may vary, especially in entailment relations such as skeptical or credulous $c$-inference. The most direct relationship to explore is the comparison of $\kappa$-entailment for classical DL statements with $\vDash^{opt}_c$ and $\vDash^{opt}_p$.

\begin{proposition}\label{proposition:opt-c-entailment-expressible-as-kappa-entailment}
    For a given WKB $\KB_\omega$ and any GCI or ABox statement $\tau$ we have that $\KB_\omega\vDash ^{opt}_c \tau$ iff $\KB\dentails^{\kappa_\omega}\tau$, where $\KB$ is the quantified or open translated knowledge base. Similarly, given a defeasible knowledge base $\KB=(\T,\D,\A)$, a $c$-representation $\kappa$ and a classical GCI or ABox statement $\tau$, we have that $\KB\dentails^{\kappa}\tau$ iff $\KB^\kappa_\omega\vDash ^{opt}_c \tau$.
\end{proposition}

On the other hand, we are able to express $\vDash^{opt}_p$ as a negation of $\kappa$-entailment for the translated $c$-representation.

\begin{proposition}\label{proposition:opt-p-entailment-expressible-as-negative-kappa-entailment}
    Let $\KB_\omega$ be a WKB. Then:
    \begin{itemize}
        \item If $\cass{a}{A}$ is a concept assertion, then $\KB_\omega \vDash ^{opt}_p \cass{a}{A}$ iff $\KB\ndentails^{\kappa_\omega}\cass{a}{\neg A}$.
        \item If $C\subs D$ is a GCI then $\KB_\omega \vDash ^{opt}_p  C\subs D$ iff $\KB\ndentails^{\kappa_\omega} \cass{a}{C\dlAnd \neg D}$ for all $a\in \U$.
    \end{itemize}

    where $\KB$ is the quantified or open translated knowledge base.
\end{proposition}

This result should be unsurprising since $\KB_\omega\vDash ^{opt}_c \tau$ and $\KB\dentails^{\kappa}\tau$ results when $\tau$ holds for all interpretations with minimal cost or rank in the cost-based semantics and $c$-representation respectively. Since the translated interpretations have the same relative cost of interpretations, those with minimal cost are the same interpretations once translated. This relationship is not as clear when it comes to $\kappa$-entailment for DCIs and the relations $\vDash ^k_p$ and  $\vDash ^k_c$, especially since there is no way to fix a specific numerical threshold for rankings in $c$-representation based entailment, such as $k$ is fixed for  $\vDash ^k_p$ and  $\vDash ^k_c$. However, we are able to express entailment for DCIs in terms of $\vDash^k_p$ for the WKB translation of a $c$-representation.

\begin{proposition}\label{proposition:dsubs-expressible-as-some-k-p-entailment}
   For a defeasible knowledge base $\KB$ and a $c$-representation $\kappa$ which is a model of $\KB$, we have that $\KB\dentails^\kappa C\dsubs D$ iff there exists some $k\in \mathbb{N}$ such that $\KB^\kappa_\omega \vDash^k_p\cass{a}{C\dlAnd D}$ for some $a\in \U$ and  $\KB^\kappa_\omega \nvDash^k_p\cass{b}{C\dlAnd \neg D}$ for all $b\in \U$.
\end{proposition}

When we consider entailment relations from defeasible reasoning which consider more than one $c$-representation, it is clear that this does not correspond to one specific WKB, but rather a class of $c$-compatible WKBs where the ABox and TBox remain the constant, but the weighting function $\omega$ varies. This is made precise in the following results. In the following proposition, for a defeasible knowledge base $\KB$, we define the class of WKBs considered by {$\text{WKB}_\KB:=\{\KB^\kappa_\omega\mid \kappa\text{ is a $c$-representation which is a model of }\KB\}$}.

\begin{proposition}\label{proposition:classical-statements-cred-and-skep-inference}
  Let $\KB=(\T,\A,\D)$ be a defeasible knowledge base. Then, if $\tau$ is a classical $\ALCO$ statement\footnote{That is, $\tau$ is an assertion or a GCI.} we have that $\KB\dentails^c_{sk} \tau$ iff $\KB^\kappa_\omega\vDash^{opt}_c\tau$ for all $\KB^\kappa_\omega\in \text{WKB}_\KB$.

    Similarly, $\KB\dentails^c_{cr} \tau$ iff $\KB^\kappa_\omega\vDash^{opt}_c\tau$ for some $\KB^\kappa_\omega\in \text{WKB}_\KB$.
\end{proposition}

We are also able to show that, under certain conditions,  $\dentails^c_{sk}$ is a stronger notion of $\vDash^{opt}_c$ once we translate a given WKB into a defeasible knowledge base.

\begin{corollary}\label{corol:skept-c-inf-stronger-than-opt-c}
    For any $c$-compatible WKB $\KB_\omega$, and any classical $\ALCO$ statement $\tau$, $\KB\dentails^c_{sk} \tau$  implies that $\KB_\omega \vDash^{opt}_c\tau$,
    where $\KB$ is the defeasible knowledge base in the open $c$-representation translation of $\KB_\omega$.
\end{corollary}

We are therefore able to rephrase certain entailment relations in defeasible reasoning in terms of cost-based semantics, as well as rephrasing certain entailment relations in cost-based semantics in terms of $c$-representations, although these translations may not cover the full expressivity of the original relation. This does not only point to the potential for more unification between inconsistency tolerant semantics and defeasible reasoning in DLs, but it also allows each approach to possibly inherit techniques and results from the other. For example, defeasible reasoning using $c$-representations stands to benefit from established complexity results in cost-based semantics \cite{BienvenuJeanCostBased}, while work done in computing $c$-representations, such as reducing skeptical $c$-inference to a Constraint Satisfaction Problem \cite{BeierleEichhornKern-Isberner:skepticaljournal} may be applicable to cost-based semantics. 

\section{Related Work}\label{sect_related_work}
The current paper is based on the work of Bienvenu et al. \cite{BienvenuJeanCostBased}, who introduce cost-based semantics for query answering over inconsistent knowledge bases, and on the work of Kern-Isberner and others \cite{DBLP:conf/ecai/Kern-IsbernerT12,DBLP:conf/nmr/HahnKM24}, who define ranking functions and $c$-representations for conditional FOL. Defeasible concept inclusions in DLs have been considered using preferential semantics in $\ALC$ by Giordano et al. \cite{Giordano13,GiordanoEtAl2015} and Britz et al. \cite{BritzEtAl21}, in the more expressive DL $\mathcal{SROIQ}$ by Britz and Varzinczak \cite{BritzVarz17}, and in less expressive DLs by Pensel and Turhan \cite{PenselTurhan18-defeasibleEL} and by Casini and others \cite{CasiniSM19}, although these works do not use ranking-function based semantics. Ranking functions as a semantics for $\ALC$ has been considered by Hahn et al. \cite{DBLP:conf/nmr/HahnKM24}, while a ranking semantics for a Datalog style restriction of first order logic has been considered by Casini et al. \cite{CasiniMeyerPJ:klmrfol}.

Cost-based semantics are related to repair-based semantics considered by Bienvenu et al. \cite{BienvenuEtAl2014,BienvenuEtAl2016} and Lembo et al. \cite{LemboEtAl2010}, who approach inconsistency in DL knowledge bases by making principled alterations to the knowledge base in order to obtain a consistent knowledge base. A similar approach to cost-based semantics is given in the framework of existential rules by Eiter et al. \cite{EiterLP16softrules}, who consider soft databases and soft programs, where models with minimal instances of broken rules in a given soft program are considered for query-answering purposes.

\section{Conclusions}\label{sect_conclusion}

In this paper, we provide a comparison between the frameworks of cost-based semantics for inconsistency-tolerant query answering in the DL $\ALCO$, and the ranking function based semantics of  $c$-representations for defeasible reasoning in First-Order Logic. In doing so, we highlight the similarities between both frameworks and suggest ways in which certain aspects can be expressed by the other under certain conditions. In particular, we show that for a given $c$-representation $\kappa$ for a defeasible knowledge base $\KB$, we can define a weighted knowledge based $\KB_\omega$ such that $cost_{\KB_\omega}(\I)<cost_{\KB_\omega}(\I')$ iff. $\kappa(\I)<\kappa(\I')$, and show that when certain conditions are satisfied, the converse is possible. Moreover, we consider the resulting links between entailment in each framework and show that certain entailment relations which are defined for one framework can be expressed in terms of entailment relations defined in the other. Overall, the goal of this paper is to provide a technically precise means to understand one framework in terms of the other, and provide a starting point for unifying both approaches from the query-answering and defeasible reasoning communities, where desirable. This has the potential to benefit both frameworks, since under certain conditions they stand to inherit methodologies and results from the other framework, as well as increase the expressivity and scope of applicability for both frameworks. 

As such, immediate future work to consider is to what extent results which hold in one framework can be applied to the other. In particular, it is worth investigating the complexity results given in cost-based semantics \cite{BienvenuJeanCostBased} as a means to determine complexity bounds for defeasible reasoning in DLs. On the other hand, it would be worth investigating methods used for determining relevant ranking functions within defeasible reasoning, such as the Constraint Satisfaction problems considered in \cite{BeierleEichhornKern-Isberner:skepticaljournal}, as a means to algorithmically determine cost functions within weighted knowledge bases, rather than requiring such cost functions to be declared. In order to further unify the field, a more detailed comparison between cost-based semantics and other semantics for defeasible descriptions logics, such as the more widely used preferential interpretations \cite{BritzEtAl21}, could be considered.

\begin{acknowledgments}
This work is based on the research supported in part by the National Research Foundation of South Africa (REFERENCE NO: SAI240823262612). The work of Giovanni Casini and Thomas Meyer has  been partially supported by the H2020 STARWARS Project (GA No. 101086252), action HORIZON TMA MSCA Staff Exchanges. 
\end{acknowledgments}

\bibliography{DLbib}

\newpage
\appendix
\section{Proofs for Sections \ref{sect_costsemantics} and \ref{sect_c_repres}}

\noindent\textbf{Proposition \ref{proposition:cost-based-entailments-monotonic-or-not}} \textit{For any $k\in\mathbb{N}$, the entailment $\vDash^k_{c}$ is monotonic, while $\vDash^k_{p}$, $\vDash^{opt}_{c}$ and $\vDash^{opt}_{p}$ are non-monotonic.}

\begin{proof}
    Let $\KB_\omega$ be an arbitrary WKB. We see that  $\vDash^k_{c}$ is monotonic for the following reasons. Firstly, if we add a new statement $
\tau$ to the knowledge base to obtain $\KB'_{\omega'}$, then, for any interpretation $\I$, we either have $\I\Vdash \tau$ or $\I \nVdash \tau$. Since $\omega(\tau)\geq 0$, we then have if  $\I\Vdash \tau$ then $cost_{\KB'_{\omega'}}(\I)=cost_{\KB_\omega}(\I)$, and if $\I\nVdash \tau$ then $cost_{\KB'_{\omega'}}(\I)\geq cost_{\KB_\omega}(\I)$. In either case we have $cost_{\KB'_{\omega'}}(\I)\geq cost_{\KB_\omega}(\I)$. Then, if $\KB_\omega\vDash^k_{c} q$, we have that $\I\Vdash q$ for all $\I$ with $cost_{\KB_\omega}(\I)\leq k$. But then, for any $\I'$ with $cost_{\KB'_{\omega'}}(\I')\leq k$, we have that $cost_{\KB_\omega}(\I')\leq cost_{\KB'_{\omega'}}(\I')$. Then since $\KB_\omega\vDash^k_{c} q$, we must have $\I'\Vdash q$. Then, by definition $\KB'_{\omega'}\vDash^k_{c} q$ and so $\vDash^k_{c}$ is monotonic.\\

We provide a simple counterexample to show that $\vDash^k_{p}$, $\vDash^{opt}_{c}$ and $\vDash^{opt}_{p}$ are non-monotonic. Consider a signature with no role names, one concept name $\NC=\{A\}$ and a single individual name $\NI=\{a\}$. Then let the WKB $\KB_\omega$ be defined by $\A=\{A(a)\}$ and $\T=\emptyset$ where $\omega(A(a))=1$. Then we consider two interpretations. The first, $\I_1$ has $A^{\I_1}=\{a\}$, and the second $\I_2$ has $A^{\I_2}=\emptyset$. Clearly, $cost_{\KB_\omega}(\I_1)=0$ and $cost_{\KB_\omega}(\I_2)=1$. Therefore $\KB_\omega\vDash^1_{p} A(a)$, and $\KB_\omega\vDash^{opt}_{p} A(a)$. Note here since the optimal cost is bound by zero $\I_1$ must be an interpretation with optimal cost. Moreover, every interpretation $\I$ with cost $0$ must have that $\I\Vdash A(a)$, and so $\KB_\omega\vDash^{opt}_{c} A(a)$. However, if we add to our knoweldge base the ABox axiom $\neg A(a)$, such that $\omega(\neg A(a))=2$, then we observe the following.  $cost_{\KB'_{\omega'}}(\I_1)=2$ and $cost_{\KB'_{\omega'}}(\I_2)=1$, and furthermore any interpretation $\I$ such that $\I\Vdash A(a)$ has $cost_{\KB'_{\omega'}}(\I)\geq2$. Hence, $\KB'_{\omega'}\nvDash^1_{p} A(a)$. Moreover, the knowledge base is no longer classically satisfiable and so the optimal cost is $1$. Therefore $\KB'_{\omega'}\nvDash^1_{p} A(a)$ implies that $\KB'_{\omega'}\nvDash^{opt}_{p} A(a)$. Lastly,  $\KB'_{\omega'}\nvDash^{opt}_{c} A(a)$ since for example $\I_2\Vdash\neg A(a)$ and $\I_2$ has optimal cost. 
\end{proof}

\noindent\textbf{Proposition \ref{proposition:express-quantified-DCI-as-normal_DCIs}. }    \textit{$\kappa\Vdash C\dsubs^\forall D \text{ iff }\kappa\Vdash \{a\}\dlAnd C\dsubs D \text{ for all }a\in U_\Sigma$.}

\begin{proof}
   $\kappa\Vdash C\dsubs^\forall D$, iff $\kappa(\cass{a}{C\dlAnd D})<\kappa (\cass{a}{C\dlAnd \neg D})$ for all $a\in \U$. 

   On the other hand $\kappa\Vdash \{a\}\dlAnd C\dsubs D \text{ for all }a\in U_\Sigma$ iff $\kappa(\{a\}\dlAnd C\dlAnd D)< \kappa(\{a\}\dlAnd C\dlAnd \neg D)$ for all $a\in \U$. Note here that we can exclude the second condition in Definition \ref{definition:satisfaction-of-open-DCIs}, since the only possible representative for $\{a\}\dlAnd C\dsubs D$ or $\{a\}\dlAnd C\dsubs \neg D$ is $a$, since it is the only member of the Herbrand universe which can be included in  concepts $\{a\}\dlAnd C\dlAnd D$ or $\{a\}\dlAnd C\dlAnd \neg D$. Therefore if $\kappa(\{a\}\dlAnd C\dlAnd D)= \kappa(\{a\}\dlAnd C\dlAnd \neg D)$ we have $\kappa(\{a\}\dlAnd C\dlAnd D)= \kappa(\cass{a}{ C\dlAnd D)}$ and $\kappa(\{a\}\dlAnd C\dlAnd \neg D)= \kappa(\cass{a}{ C\dlAnd \neg D)}$ and so the second case in Definition \ref{definition:satisfaction-of-open-DCIs} cannot hold. From the above, it also holds that  $\kappa(\{a\}\dlAnd C\dlAnd D)< \kappa(\{a\}\dlAnd C\dlAnd \neg D)$ is equivalent to $\kappa(\cass{a}{C\dlAnd D})<\kappa (\cass{a}{C\dlAnd \neg D})$. Then, since this holds for all $a\in\U$, this is equivalent to    $\kappa\Vdash C\dsubs^\forall D$.
\end{proof}

\section{Proofs for Section \ref{sect_comparison_main}}

\noindent\textbf{Proposition \ref{proposition:from-c-rep-to}.} \textit{For a given $\ALC\mathcal{O}$ $c$-representation $\kappa$ we have
    \[\textit{cost}_{\KB^\kappa_\omega}(\I)=\kappa(\I)-\kappa_0\]  for every interpretation $\I\in\Omega_\N$.}

\begin{proof}
    Recall that 
    \begin{displaymath}
    \textit{cost}_{\KB^\kappa_\omega}(\I)=\sum\limits_{\tau\in\TBox}\omega(\tau)|\textit{vio}_{\tau}(\I)|+\sum\limits_{\alpha\in\textit{vio}_\ABox(\I)}\omega(\alpha)
\end{displaymath}

In the case where $\I\not\sat  \tau$ for some $\tau\in \ABox$ we have by construction that $\textit{cost}_{\KB^\kappa_\omega}(\I)=\infty$. Moreover, we have that $\I\not\sat  \tau$ implies that $\kappa(\I)=\infty$ by definition. Then since $\kappa_0$ is finite we have that $\textit{cost}_{\KB^\kappa_\omega}(\I)=\infty=\infty-\kappa_0=\kappa(\I)-\kappa_0$.

If $v\not\sat  \tau$ for some $\tau\in \T$ then similarly both $\textit{cost}_{\KB^\kappa_\omega}(\I)$ and $\kappa(v)$ are infinite and so $\textit{cost}_{\KB^\kappa_\omega}(\I)=\infty=\kappa(\I)-\kappa_0$.

We then consider the case where $\I\sat \tau$ for all $\tau\in\ABox\cup \TBox$. That is, when the rules that are violated only have a finite impact. We note that in this case, for any $C_i\subs D_i\in\T^*$, then $\textit{vio}_{C_i\subs D_i}(\I)=(C_i\dlAnd\neg D_i)^\I$ therefore $|\textit{vio}_{C_i\subs D_i}(\I)|=|(C_i\dlAnd\neg D_i)^\I|=f_i(\I)$ by definition.

Moreover, $\eta_i=\omega(C_i\subs D_i)$ and so:

\begin{displaymath}
    \textit{cost}_{\KB^\kappa_\omega}(\I)=\sum\limits_{\tau\in\T^*}\omega(\tau)|\textit{vio}_{\tau}(\I)| = \sum\limits_{i=1}^n\omega(C_i\subs D_i)|\textit{vio}_{C_i\subs D_i}(\I)|= \sum\limits_{i=1}^n\eta_i f_i(v)=\kappa(\I)-\kappa_0
\end{displaymath}
\end{proof}

\noindent\textbf{Proposition \ref{proposition:open-quantified-c-rep-translation-gives-same-ranking}.}\textit{ For any WKB $\KB_\omega=((\TBox,\ABox),\omega)$, let $\kappa_1$ be the function generated by the quantified c-representation translation and let $\kappa_2$ be the function generated by the open c-representation translation. Then $\kappa_1=\kappa_2$.
}

\begin{proof}
    By definition $\kappa_{1,0}=\kappa_{2,0}$, and so it is sufficient to show for any $\I\in\Omega_\N$ that 
    \[\sum_{\tau\in\D^1}f^1_\tau(\I)\eta^1_\tau=\sum_{\tau\in\D^2}f^2_\tau(\I)\eta^2_\tau\]

    where $\D^1$, $f^1_\tau$ and $\eta^1_\tau$ is defined for the quantified translation as in Definition \ref{definition:quantified-c-rep-translation} and $\D^2$, $f^2_\tau$ and $\eta^2_\tau$ is defined for the open translation as in Definition \ref{definition:open-c-rep-translation}.
We first note that $\D^1=\{\{a\}\dlAnd C\dsubs D\mid a\in\U, C\dsubs D\in \D^2\}$. We then denote for each $C\dsubs D\in\D^2$ the subset of $\D^1$ defined by $\D^1(C\dsubs D)=\{\{a\}\dlAnd C\dsubs D\mid a\in\U\}$.

Then, if  $C\dsubs D\in \D^2$ and $a\in (C\dlAnd \neg D)^\I$ we have that $(\{a\}\dlAnd C\dlAnd D)^\I=\{a\}$. This is equivalent to $\I\nVdash \{a\}\dlAnd C\subs D$. Therefore, for $\tau=C\dsubs D\in\D^2$, we have $f^2_\tau(\I)=|\{a\in \U \mid \I\nVdash \{a\}\dlAnd C\subs D\}|=|\{\tau'\in\D^1(C\dsubs D)\mid \I\nVdash \{a\}\dlAnd C\subs D\text{ where } \tau'=\{a\}\dlAnd C\dsubs D \}|$. 

Then note that since for any $\tau'=\{a\}\dlAnd C\dsubs D\in \D^1(C\dsubs D)$, we have by definition that $\eta^2_{\tau'}=\eta^1_\tau$ where $\tau=C\dsubs D$, and moreover, $f^1_{\tau'}(\I)=1$ iff $\I\nVdash \{a\}\dlAnd C\subs D$, and $f^1_{\tau'}=0$ otherwise. Therefore, 
\begin{displaymath}
    \sum_{\tau'\in\D^1}f^1_{\tau'}(\I)\eta^1_{\tau'}= \sum_{\tau\in\D^2}\Large(\normalsize\sum_{\tau'\in A}\eta^2_\tau)
\end{displaymath}

where $A=\{\tau'\in \D^1(\tau)\mid \I\nVdash\{a\}\dlAnd C\subs D\text{ where } \tau'=\{a\}\dlAnd C\dsubs D \}$. But then $|A|=f^2_\tau(\I)$ and so, 

\begin{displaymath}
    \sum_{\tau'\in\D^1}f^1_{\tau'}(\I)\eta^1_{\tau'}= \sum_{\tau\in\D^2}\Large(\normalsize\sum_{\tau'\in A}\eta^2_\tau)=\sum_{\tau\in\D^2}f^2_\tau(\I)\eta^2_\tau
\end{displaymath}
 This suffices to say that $\kappa_1=\kappa_2$.
\end{proof}

\noindent\textbf{Proposition \ref{proposition:WKB-to-c-rep-gives-same-ranking}}\textit{    For any WKB $\KB_\omega$ with a strict ABox, we have that 
    \[\kappa_\omega(\I)=cost_{\KB_\omega}(\I)+\kappa_0\]
    for any interpretation $v$ where $\kappa_\omega$ is the quantified or open c-representation translation of $\KB_\omega$.}

\begin{proof}
    For ease of proof, we treat $\kappa_\omega$ to be the open c-representation translation and note by Proposition \ref{proposition:open-quantified-c-rep-translation-gives-same-ranking} that the result holds for the quantified translation as well.

    We first consider the case where $cost_{\KB_\omega}(\I)=\infty$. Then, $cost_{\KB_\omega}(\I)+\kappa_0=\infty$ since $\kappa_0$ is finite. Note that this only occurs when $\I\nVdash \tau$ for some $\tau\in\A\cup\T$ such that $\omega(\tau)=\infty$. But by assumption each $\A\subseteq \tau^{-1}(\infty)$ and $\T^\infty=\{\tau\in\T\mid \omega(\tau)=\infty\}$. Therefore, if $cost_{\KB_\omega}(\I)=\infty$ then $\I\nVdash \tau$ for some $\tau\in \A\cup \T^\infty$ and by definition $\kappa_\omega(\I)=\infty=cost_{\KB_\omega}(\I)+\kappa_0$.

    If $cost_{\KB_\omega}(\I)$ is finite then
    \[cost_{\KB_\omega}(\I)=\sum\limits_{(C\subs D)\in\TBox\setminus\T^\infty}\omega(C\subs D)|(C\dlAnd\neg D)^\I|\]

    By definition, $C\subs D\in \TBox\setminus\T^\infty$ iff $C\dsubs D\in \D$. Moreover, if $\tau=C\dsubs D$, then $\eta_\tau=\omega(C\subs D)$. Lastly, $f_\tau(\I)=|(C\dlAnd\neg D)^\I|$ for all $\tau\in \D$. Therefore,

    \[\sum\limits_{(C\subs D)\in\TBox\setminus\T^\infty}\omega(C\subs D)|(C\dlAnd\neg D)^\I|=\sum_{\tau\in \D}\eta_\tau f_\tau(\I)\]

  And finally this shows that for any $\I$ with finite cost,

  \[\kappa_\omega(\I)=\sum_{\tau\in \D}\eta_\tau f_\tau(\I) +\kappa_0= cost_{\KB_\omega}(\I)+\kappa_0\]
  
and therefore the result holds.
    
\end{proof}

\noindent\textbf{Proposition \ref{proposition:every-strong-c-compatible-is-c-compatible}.}\textit{ Any strongly $c$-compatible WKB is $c$-compatible.
}

\begin{proof}
    Assume $\KB_\omega$ is strongly $c$-compatible but not $c$-compatible, in order to obtain a contradiction. Then, since $\KB_\omega$ is not $c$-compatible
    \begin{displaymath}
        \min_{a\in\mathcal{U},\I\Vdash a:A\dlAnd B}cost_{\KB_\omega}(\I)\geq \min_{a\in\mathcal{U},\I\Vdash a:A\dlAnd \neg B}cost_{\KB_\omega}(\I)
    \end{displaymath}

    Then, let $b\in\mathcal{U}$ be an element such that for some interpretation we have $\I'\Vdash b:A\dlAnd\neg B$ where $cost(\I')=\min_{a\in\mathcal{U},\I\Vdash a:A\dlAnd \neg B}cost_{\KB_\omega}(\I)$. Clearly, $cost(\I')=\min_{\I\Vdash b:A\dlAnd \neg B}cost(\I)$. Then, since $\KB_\omega$ is strongly $c$-compatible we have,

        \begin{displaymath}
        \min_{\I\Vdash b:A\dlAnd B}cost(\I)<\min_{\I\Vdash b:A\dlAnd \neg B}cost(\I)=\min_{a\in\mathcal{U},\I\Vdash a:A\dlAnd \neg B}cost_{\KB_\omega}(\I)\leq \min_{a\in\mathcal{U},\I\Vdash a:A\dlAnd B}cost_{\KB_\omega}(\I)
    \end{displaymath}

    But then,

        \begin{displaymath}
        \min_{\I\Vdash b:A\dlAnd B}cost(\I)<\min_{a\in\mathcal{U},\I\Vdash a:A\dlAnd B}cost_{\KB_\omega}(\I)
    \end{displaymath}
     which is a clear contradiction.
\end{proof}

\noindent\textbf{Theorem \ref{theorem:when-does-a-WKB-induce-a-c-representation}.}   \textit{A WKB is strongly $c$-compatible if and only if its quantified $c$-representation translation satisfies the translated knowledge base. Similarly, a  WKB is  $c$-compatible if and only if its open $c$-representation translation satisfies the translated knowledge base.}

\begin{proof}
    We begin with the quantified c-representation translation: For any $\{a\}\dlAnd C\dsubs D\in \D$ and any WKB $\KB_\omega$, let $\kappa$ be the quantified $c$-representation. We have that:

$\kappa\Vdash \{a\}\dlAnd C\dsubs D$ 

iff $\kappa(\{a\}\dlAnd C\dlAnd D)<\kappa(\{a\}\dlAnd C\dlAnd \neg D)$ 

iff $\kappa(\cass{a}{C\dlAnd D})<\kappa(\cass{a}{C\dlAnd \neg D})$ 

iff $\min_{\I\Vdash a:C\dlAnd D}\kappa(\I)< \min_{\I\Vdash a:C\dlAnd \neg D}\kappa(\I)$

iff $\min_{\I\Vdash a:C\dlAnd D}cost_{\KB_\omega}(\I)< \min_{\I\Vdash a:C\dlAnd \neg D}cost_{\KB_\omega}(\I)$ by Proposition \ref{proposition:WKB-to-c-rep-gives-same-ranking}.

Therefore, if the in the line above condition holds for all $B\subs C\in \{\tau\in\T\mid \omega(\tau)<\infty\}$, the first line holds for all $\{a\}\dlAnd C\dsubs D\in \D$. That is, the WKB is strongly c-compatible iff its quantified $c$-representation translation satisfies the translated knoweldge base.

The proof of the open translation is similar. but includes the additional notion that we consider $\kappa(C\dlAnd D)$ and $\kappa(C\dlAnd \neg D)$ instead of $\kappa(\{a\}\dlAnd C\dlAnd D)$ and $\kappa(\{a\}\dlAnd C\dlAnd \neg D)$ respectively.
\end{proof}

\noindent\textbf{Theorem \ref{theorem:open-translation-inverse}. }\textit{Let $\KB=(\T,\A,\D)$ be a defeasible knowledge base, and let $\KB_\omega$ be a $c$-compatible WKB. Then $\KB^{\kappa_\omega}_{\omega'}=\KB_\omega$, where $\kappa_\omega$ is the open $c$-representation translation of $\KB_\omega$, and $\kappa_{\omega''}=\kappa$, where $\kappa_{\omega''}$ is the open $c$-representation translation of $\KB^\kappa_{\omega''}$.}

\begin{proof}
    We start by showing that $\KB^{\kappa_\omega}_{\omega'}=\KB_\omega$. Let $\KB_\omega=((\T,\A),\omega)$ and let $\KB^{\kappa_\omega}=((\T',\A'),\omega')$. Clearly $\A=\A'$ since neither translation changes the ABox (since we assume the ABox is strict). Furthermore, the translation in Definition \ref{definition:open-c-rep-translation} converts TBox statements $C\subs D$ with finite weights into DBox statements $C\dsubs D$ while the translation in Definition \ref{definition:weight-kb-translation} translates any DBox statement $C\dsubs D$ back to the original TBox statement $C\subs D$. TBox statements with infinite weights are not changed throughout. Therefore $\T=\T'$. Clearly, $\omega(\tau)=\infty=\omega'(\tau)$ for all $\tau\in\A$. Then for any $\tau\in \T$, if $\omega(\tau)=\infty$ then $\tau\in \T^\infty$ in the knowledge base $\KB^*$ generated by the open $c$-representation translation, as given in Definition \ref{definition:open-c-rep-translation}. Then $\omega'(\tau)=\infty$ by definition and so $\omega(\tau)=\omega'(\tau)$. If $\omega(\tau)<\infty$ for $\tau=C\subs D$ then $\eta_{C\dsubs D}=\omega(\tau)$ by Definition \ref{definition:open-c-rep-translation} and $\eta_{C\dsubs D}=\omega'(\tau)$ by Definition \ref{definition:weight-kb-translation}. Therefore $\omega(\tau)=\omega'(\tau)$ for all $\tau\in \T$ and so $\omega=\omega'$.\\

    On the other hand, suppose $\kappa$ is a $c$-representation that models $\KB=(\T,\A,\D)$ and $\kappa_{\omega''}$ is the open $c$-representation translation of $\KB^\kappa_{\omega''}=((\T'',\A''),\omega'')$, where $\KB^*=(\T''^\infty,\A^*,\D^*)$ is the knoweldge base generated by translating $\KB^\kappa_{\omega''}$ as in Definition \ref{definition:open-c-rep-translation}. Then note that $\T^\infty$ consists of all those elements in $\T''$ with infinite weight, which consists of the original GCIs in $\T$. Therefore $\T''^\infty=\T$. Similarly to the last part of the proof we see that the translations preserve the SBox and so $\A^*=\A$ while the translations convert the DCIs in $\D$ into GCIs and then back once again, and so we obtain $\D=\D^*$. 

    We then show that the ranking functions are equal by considering two cases. In the first case assume $\kappa_{\omega''}(\I)=\infty$. This is true iff $\I\nVdash \tau$ for some $\tau\in\T^\infty\cup\A^*=\T\cup\A$. But then by definition we have $\kappa(\I)=\infty=\kappa_{\omega''}(\I)$.

    In the second case assume $\kappa_{\omega''}(\I)<\infty$. Then $\kappa_{\omega''}(\I)=\kappa_{\omega'',0}+\sum_{\tau\in \D^*} f_{\tau}(\I)\eta_\tau$. Then firstly note $\kappa_{\omega'',0}=-optc(\KB^\kappa_{\omega''})=--\kappa_0$, where the first equality follows from Definition \ref{definition:open-c-rep-translation} and the second follows from the fact that in Definition \ref{definition:weight-kb-translation} the minimal cost generated by $\KB^\kappa_{\omega''}$ is that which makes $\kappa(\I)=0$ by the definition of a $c$-representation and Proposition \ref{proposition:from-c-rep-to}. Then since we can express $\D^*$ by $\D^*=\D=\{C_1\dsubs D_1,...,C_n\dsubs D_n\}$ we have that $\sum_{\tau\in \D^*} f_{\tau}(\I)\eta_\tau=\sum_{i=1}^n f''_{i}(\I)\eta''_i$, where
    \begin{itemize}
        \item $\eta''_i=\omega''(C_i\subs D_i)=\eta_i$ by definition of the translations.
        \item $f''_i(\I)=|(C\dlAnd \neg D)^\I|=f_i(\I)$ by definition of the translations.
    \end{itemize}

    Therefore $\kappa_{\omega''}(\I)=\sum_{i=1}^n f_{i}(\I)\eta_i=\kappa(\I)$ in the case where $\kappa_{\omega''}(\I)$ is finite. Thus, $\kappa_{\omega''}=\kappa$.
\end{proof}

\noindent\textbf{Proposition \ref{proposition:abox-translation-preserves-costs}. }\textit{  For any WKB $\KB_\omega$, we have that 
    \[cost_{\KB_\omega}(\I)=cost_{\KB'_{\omega'}}(\I)\]
    for all $\I\in\Omega_\N$, where $\KB'_{\omega'}$ is the strict ABox translation of $\KB_\omega$.}

\begin{proof}
    Let $\KB_\omega=((\T,\A),\omega)$ be a WKB and let  $\KB'_{\omega'}$ be the strict Abox translation of $\KB_\omega((\T',\A'),\omega')$.
    For any $\I \in\Omega_\N$

\begin{displaymath}
cost_{\KB_\omega}(\I)=\sum\limits_{\tau\in\TBox}\omega(\tau)|\textit{vio}_{\tau}(\I)|+\sum\limits_{\alpha\in\textit{vio}_\ABox(\I)}\omega(\alpha)
\end{displaymath}
    and 
    \begin{displaymath}
cost_{\KB'_{\omega'}}(\I)=\sum\limits_{\tau\in\TBox'}\omega'(\tau)|\textit{vio}_{\tau}(\I)|+\sum\limits_{\alpha\in\textit{vio}_{\ABox'}(\I)}\omega'(\alpha)
\end{displaymath}

    Then note that, since $\omega(\tau)=\omega'(\tau)$ for all $\tau\in\A'\cup\T$. We have that 

\begin{align*}          &\sum\limits_{\tau\in\TBox'}\omega'(\tau)|\textit{vio}_{\tau}(\I)|+\sum\limits_{\alpha\in\textit{vio}_{\ABox'}(\I)}\omega'(\alpha)\\
=&\sum\limits_{\tau\in\TBox}\omega(\tau)|\textit{vio}_{\tau}(\I)|+ \sum\limits_{\tau\in\TBox'\setminus\T}\omega'(\tau)|\textit{vio}_{\tau}(\I)|+\sum\limits_{\alpha\in\textit{vio}_{\ABox'}(\I)}\omega'(\alpha).
\end{align*}

Similarly, 

\begin{align*}
&\sum\limits_{\tau\in\TBox}\omega(\tau)|\textit{vio}_{\tau}(\I)|+\sum\limits_{\alpha\in\textit{vio}_\ABox(\I)}\omega(\alpha)\\
=&\sum\limits_{\tau\in\TBox}\omega(\tau)|\textit{vio}_{\tau}(\I)|+\sum\limits_{\alpha\in\textit{vio}_{\ABox\setminus \ABox'}(\I)}\omega(\alpha)+\sum\limits_{\alpha\in\textit{vio}_{\ABox'}(\I)}\omega'(\alpha).
\end{align*}

Therefore it is sufficient to show that \[\sum\limits_{\tau\in\TBox'\setminus\T}\omega'(\tau)|\textit{vio}_{\tau}(\I)|=\sum\limits_{\alpha\in\textit{vio}_{\ABox\setminus \ABox'}(\I)}\omega(\alpha).\]

Then note that $\textit{vio}_{\ABox\setminus \ABox'}(\I)=\{\alpha\in \ABox\setminus \ABox'\mid \I\nVdash \alpha\}$. But then for any $\cass{a}{A}\in\A\setminus\A'$ there exists $\{a\}\subs A\in \T'\setminus \T$. Moreover, whenever $\I\nVdash \cass{a}{A}$ we have that $a\notin A^\I$. Equivalently $\{a\}^\I\nsubseteq A^\I$ or $\I\nVdash \{a\}\subs A$. Then note that for any such $\{a\}\subs A$ where $\I\nVdash \{a\}\subs A$, $vio_{\{a\}\subs A}(\I)=\{a\}$, and so $|vio_{\{a\}\subs A}(\I)|=1$. Therefore, there is some GCI $\{a\}\subs A\in \T'\setminus\T$ which is violated by exactly one element in $\I$ for every $\cass{a}{A}\in \textit{vio}_{\ABox\setminus \ABox'}(\I)$. Moreover, $\omega'(\{a\}\subs A)=\omega(\cass{a}{A})$ by definition. And so \[\sum\limits_{\tau\in\TBox'\setminus\T}\omega'(\tau)|\textit{vio}_{\tau}(\I)|=\sum\limits_{\alpha\in\textit{vio}_{\ABox\setminus \ABox'}(\I)}\omega(\alpha).\]
\end{proof}

\noindent\textbf{Proposition \ref{proposition:opt-c-entailment-expressible-as-kappa-entailment}. }\textit{For a given WKB $\KB_\omega$ and any GCI or ABox statement $\tau$ we have that $\KB_\omega\vDash ^{opt}_c \tau$ iff $\KB\dentails^{\kappa_\omega}\tau$, where $\KB$ is the translated knowledge base. Similarly, given a defeasible knowledge base $\KB=(\T,\A,\D)$, a $c$-representation $\kappa$ and a classical GCI or ABox statement $\tau$, we have that $\KB\dentails^{\kappa}\tau$ iff $\KB^\kappa_\omega\vDash ^{opt}_c \tau$.}

\begin{proof}
    For the first part of the proof let  $\KB_\omega$ be a (strongly $c$-compatible) WKB, and let $\tau$ be a classical GCI or concept/role assertion. Then, if $\kappa_\omega$ is the open or quantified $c$-representation translation of $\KB_\omega$\footnote{We can examine both open and quantified translations simultaneously since, by Proposition \ref{proposition:open-quantified-c-rep-translation-gives-same-ranking}, $\kappa_1^{-1}(0)=\kappa_2^{-1}(0)$ where $\kappa_1$ is the quantified translation, and $\kappa_2$ is the open translation.}, we have that $\KB\dentails^{\kappa_\omega}\tau$ iff $\I\Vdash \tau$ for all $\I\in\kappa_\omega^{-1}(0)$. Then by Proposition \ref{proposition:WKB-to-c-rep-gives-same-ranking} we have that $cost_{\KB_\omega}(\I)=optc(\KB)$ iff $\kappa_\omega(\I)=0$. Therefore $\KB_\omega\vDash ^{opt}_c \tau$ iff, $\I\Vdash \tau$ for all $\I$ such that $cost_{\KB_\omega}(\I)=optc(\KB)$, iff $\I\Vdash \tau$ for all $\I\in \kappa_\omega^{-1}(0)$, iff $\KB\dentails^{\kappa_\omega} \tau$.

    The second part of the Proposition is proved using a similar argument and the fact that $cost_{\KB^\kappa_\omega}(\I)=optc(\KB^\kappa_\omega)$ iff $\kappa(\I)=0$, which is a corollary of Proposition \ref{proposition:from-c-rep-to}.
\end{proof}

\noindent\textbf{Proposition \ref{proposition:opt-p-entailment-expressible-as-negative-kappa-entailment}. }
\textit{Let $\KB_\omega$ be a WKB. Then:}
    \begin{itemize}
        \item \textit{If $\cass{a}{A}$ is a concept assertion, then $\KB_\omega \vDash ^{opt}_p \cass{a}{A}$ iff $\KB\ndentails^{\kappa_\omega}\cass{a}{\neg A}$.}
        \item \textit{If $C\subs D$ is a GCI then $\KB_\omega \vDash ^{opt}_p  C\subs D$ iff$\KB\ndentails^{\kappa_\omega} \cass{a}{C\dlAnd \neg D}$ for all $a\in \U$.}
    \end{itemize}
\textit{where $\KB$ is the quantified or open translated knowledge base.}

\begin{proof}
    \begin{itemize}
        \item For concept assertions, we note that each of the following statements are equivalent:
        \begin{itemize}
            \item $\KB_\omega \vDash ^{opt}_p \cass{a}{A}$;
            \item $\I\Vdash \cass{a}{A}$ for some $\I$ such that $cost_{\KB_\omega}(\I)=optc(\KB)$;
            \item $\I\Vdash \cass{a}{A}$ for some $\I$ such that $\kappa_\omega(\I)=0$ by Proposition \ref{proposition:WKB-to-c-rep-gives-same-ranking};
            \item $a\in A^\I$ for some $\I$ such that $\kappa_\omega(\I)=0$;
            \item $a\notin (\neg A)^\I$ for some $\I$ such that $\kappa_\omega(\I)=0$;
            \item $\I\nVdash \cass{a}{\neg A}$ for some $\I$ such that $\kappa_\omega(\I)=0$;
            \item $\KB\ndentails^{\kappa_\omega} \cass{a}{\neg A}$.
        \end{itemize}
        \item Note for GCIs, that the following statements are equivalent:
        \begin{itemize}
          \item $\KB_\omega \vDash ^{opt}_p C\subs D$;
            \item $\I\Vdash C\subs D$ for some $\I$ such that $cost_{\KB_\omega}(\I)=optc(\KB)$;
            \item $\I\Vdash C\subs D$ for some $\I$ such that $\kappa_\omega(\I)=0$ by Proposition \ref{proposition:WKB-to-c-rep-gives-same-ranking};
            \item $C^I\subseteq D^\I$ for some $\I$ such that $\kappa_\omega(\I)=0$;
            \item $C^\I\cap(\Delta^\I\setminus D^\I)=(C\dlAnd \neg D)^\I=\emptyset$ for some $\I$ such that $\kappa_\omega(\I)=0$;
            \item For all $a\in \U$, $\I\nVdash \cass{a}{C\dlAnd \neg D}$ for some $\I$ such that $\kappa_\omega(\I)=0$;
            \item $\KB\ndentails^{\kappa_\omega} \cass{a}{C\dlAnd \neg D}$ for all $a\in \U$.
        \end{itemize}
    \end{itemize}
\end{proof}

\noindent\textbf{Proposition \ref{proposition:dsubs-expressible-as-some-k-p-entailment}. }\textit{For a defeasible knowledge base $\KB$ and a $c$-representation $\kappa$ which is a model of $\KB$, we have that $\KB\dentails^\kappa C\dsubs D$ iff there exists some $k\in \mathbb{N}$ such that $\KB^\kappa_\omega \vDash^k_p\cass{a}{C\dlAnd D}$ for some $a\in \U$ and  $\KB^\kappa_\omega \nvDash^k_p\cass{b}{C\dlAnd \neg D}$ for all $b\in \U$.}

\begin{proof}
    $(\impliedby):$ Suppose for some defeasible knowledge base $\KB$ and some $c$-representation $\kappa$ we have that the translated WKB $\KB^\kappa_\omega$ satisfies the conditions that there exists some $k\in \mathbb{N}$ such that $\KB^\kappa_\omega \vDash^k_p\cass{a}{C\dlAnd D}$ for some $a\in \U$ and  $\KB^\kappa_\omega \nvDash^k_p\cass{b}{C\dlAnd \neg D}$ for all $b\in \U$. This is equivalent to saying that $\I\Vdash \cass{a}{C\dlAnd D}$ for some $\I$ eith $cost_{\KB^\kappa_\omega}\leq k$, and there is no $\I\in\Omega_\N$ and $b\in\U$ such that $\I\Vdash\cass{b}{C\dlAnd \neg D}$ and $cost_{\KB^\kappa_\omega}(\I)\leq k$. Then, 
\begin{displaymath}
    k<\min_{b\in\U, \I\Vdash b:C\dlAnd\neg D} cost_{\KB^\kappa_\omega}(\I)
\end{displaymath}

and 

\begin{displaymath}
    \min_{c\in\U, \I\Vdash c:C\dlAnd D} cost_{\KB^\kappa_\omega}(\I)\leq \min_{\I\Vdash a:C\dlAnd D} cost_{\KB^\kappa_\omega}(\I)\leq k
\end{displaymath}

and therefore

\begin{displaymath}
    \min_{b\in\U, \I\Vdash b:C\dlAnd D} cost_{\KB^\kappa_\omega}(\I)< \min_{b\in\U, \I\Vdash b:C\dlAnd\neg D} cost_{\KB^\kappa_\omega}(\I)
\end{displaymath}

But then since by Proposition \ref{proposition:from-c-rep-to} $cost_{\KB^\kappa_\omega}(\I)<cost_{\KB^\kappa_\omega}(\I')$ iff $\kappa(\I)<\kappa(\I')$, we have that 

\begin{displaymath}
    \min_{b\in\U, \I\Vdash b:C\dlAnd D} \kappa(\I)< \min_{b\in\U, \I\Vdash b:C\dlAnd\neg D} \kappa(\I)
\end{displaymath}
 which is equivalent to $\kappa(C\dlAnd D)<\kappa(C\dlAnd \neg D)$ and therefore $\kappa\Vdash C\dsubs D$.

 $(\implies):$ Assume $\kappa\Vdash C\dsubs D$. Then $\kappa(C\dlAnd D)<\kappa(C\dlAnd \neg D)$, and so 
 
 \begin{displaymath}
    \min_{b\in\U, \I\Vdash b:C\dlAnd D} \kappa(\I)< \min_{b\in\U, \I\Vdash b:C\dlAnd\neg D} \kappa(\I)
\end{displaymath}

which, as above is equivalent to 

\begin{displaymath}
    \min_{b\in\U, \I\Vdash b:C\dlAnd D} cost_{\KB^\kappa_\omega}(\I)< \min_{b\in\U, \I\Vdash b:C\dlAnd\neg D} cost_{\KB^\kappa_\omega}(\I).
\end{displaymath}

Then we choose $k=\min_{b\in\U, \I\Vdash b:C\dlAnd D} cost_{\KB^\kappa_\omega}(\I)$, and by definition there exists some $a\in \U$ such that $\KB^\kappa_{\omega}\vDash^k_p \cass{a}{C\dlAnd D}$. However since $k < \min_{b\in\U, \I\Vdash b:C\dlAnd\neg D} \kappa(\I)$ there is no $b\in \U$ such that $\I\vDash^k_p \cass{b}{C\dlAnd \neg D}$, and we are done.
\end{proof}

\noindent\textbf{Proposition \ref{proposition:classical-statements-cred-and-skep-inference}. }\textit{  Let $\KB=(\T,\A,\D)$ be a defeasible knowledge base. Then, if $\tau$ is a classical $\ALCO$ statement\footnote{That is, $\tau$ is an assertion or a GCI.} we have that $\KB\dentails^c_{sk} \tau$ iff $\KB^\kappa_\omega\vDash^{opt}_c\tau$ for all $\KB^\kappa_\omega\in \text{WKB}_\KB$.}

\textit{Similarly, $\KB\dentails^c_{cr} \tau$ iff $\KB^\kappa_\omega\vDash^{opt}_c\tau$ for some $\KB^\kappa_\omega\in \text{WKB}_\KB$.}

\begin{proof}
    Note the following:
     $\KB\dentails^c_{sk} \tau$

     $\Longleftrightarrow \kappa\Vdash \tau$ for all $c$-representations $\kappa$ which are models of $\KB$;

     $\Longleftrightarrow \KB^\kappa_\omega\vDash^{opt}_c\tau$ for all $c$-representations $\kappa$ which are models of $\KB$, by Proposition \ref{proposition:opt-c-entailment-expressible-as-kappa-entailment};

     $\Longleftrightarrow \KB^\kappa_\omega\vDash^{opt}_c\tau$ for all $\KB^\kappa_\omega\in\text{WKB}_\KB$.

     The proof for credulous $c$-inference is similar.
\end{proof}

    \noindent\textbf{Corollary \ref{corol:skept-c-inf-stronger-than-opt-c}. }\textit{For any $c$-compatible WKB $\KB_\omega$, we have that for any classical $\ALCO$ statement $\tau$:
    \[\KB\dentails^c_{sk} \tau\implies \KB_\omega \vDash^{opt}_c\tau \]
    where $\KB$ is the defeasible knowledge base in the open  $c$-representation translation of $\KB_\omega$.}

    \begin{proof}
        If $\KB_\omega$ is $c$-compatible then for the translated knowledge base $\KB$, by Theorem \ref{theorem:when-does-a-WKB-induce-a-c-representation} the derived $c$-representation $\kappa_\omega$ is a well-defined model for $\KB$. Moreover, by definition we get $\KB^{\kappa_\omega}_{\omega'}\in\text{WKB}_\KB$ where $\KB^{\kappa_\omega}_{\omega'}$ is the weighted knowledge base translation of $\kappa_\omega$. But by Theorem \ref{theorem:open-translation-inverse}, $\KB^{\kappa_\omega}_{\omega'}=\KB_\omega$ and so by Proposition \ref{proposition:classical-statements-cred-and-skep-inference} we have $\KB_\omega\vDash^{opt}_c \tau$.
    \end{proof}

\end{document}